\documentclass[runningheads]{llncs}

\usepackage{eccv}

\usepackage{eccvabbrv}

\usepackage{graphicx}
\usepackage{booktabs}

\usepackage[accsupp]{axessibility}  

\usepackage[pagebackref,breaklinks,colorlinks,citecolor=eccvblue]{hyperref}

\usepackage{orcidlink}
\usepackage[misc]{ifsym}
\usepackage{multirow}
\usepackage{colortbl}
\usepackage{xcolor}
\definecolor{Gray}{gray}{0.85}
\definecolor{Gray2}{gray}{0.90}
\usepackage{amssymb}
\usepackage{pifont}
\newcommand{\cmark}{\ding{51}}%

\begin{document}

\title{AttentionHand:\\Text-driven Controllable Hand Image Generation for 3D Hand Reconstruction in the Wild} 

\titlerunning{AttentionHand}

\author{Junho Park\inst{1,2}$^*$\orcidlink{0009-0001-3474-0010} \and
Kyeongbo Kong\inst{3}$^*$\orcidlink{0000-0002-1135-7502} \and
Suk-Ju Kang\inst{1}\textsuperscript{\Letter} \orcidlink{0000-0002-4809-956X}}

\authorrunning{J. Park et al.}

\institute{
Department of Electronic Engineering, Sogang University, South Korea \and
AI Lab, CTO Division, LG Electronics, South Korea \and
Department of Electrical \& Electronics Engineering, Pusan National University, South Korea \\
\email{junho18.park@gmail.com} \email{    kbkong@pusan.ac.kr    } \email{sjkang@sogang.ac.kr} \\
\href{https://redorangeyellowy.github.io/AttentionHand/}{https://redorangeyellowy.github.io/AttentionHand/}
}

\maketitle

\def\thefootnote{*}\footnotetext{Equal contribution.}
\def\thefootnote{\Letter}\footnotetext{Corresponding author.}

\begin{abstract}
  Recently, there has been a significant amount of research conducted on 3D hand reconstruction to use various forms of human-computer interaction. However, 3D hand reconstruction in the wild is challenging due to extreme lack of in-the-wild 3D hand datasets. Especially, when hands are in complex pose such as interacting hands, the problems like appearance similarity, self-handed occclusion and depth ambiguity make it more difficult. To overcome these issues, we propose AttentionHand, a novel method for text-driven controllable hand image generation. Since AttentionHand can generate various and numerous in-the-wild hand images well-aligned with 3D hand label, we can acquire a new 3D hand dataset, and can relieve the domain gap between indoor and outdoor scenes. Our method needs easy-to-use four modalities (i.e, an RGB image, a hand mesh image from 3D label, a bounding box, and a text prompt). These modalities are embedded into the latent space by the encoding phase. Then, through the text attention stage, hand-related tokens from the given text prompt are attended to highlight hand-related regions of the latent embedding. After the highlighted embedding is fed to the visual attention stage, hand-related regions in the embedding are attended by conditioning global and local hand mesh images with the diffusion-based pipeline. In the decoding phase, the final feature is decoded to new hand images, which are well-aligned with the given hand mesh image and text prompt. As a result, AttentionHand achieved state-of-the-art among text-to-hand image generation models, and the performance of 3D hand mesh reconstruction was improved by additionally training with hand images generated by AttentionHand.
\end{abstract}

\section{Introduction}
\begin{figure*}[t!]
\centering
\begin{center}
\includegraphics[width=\linewidth]{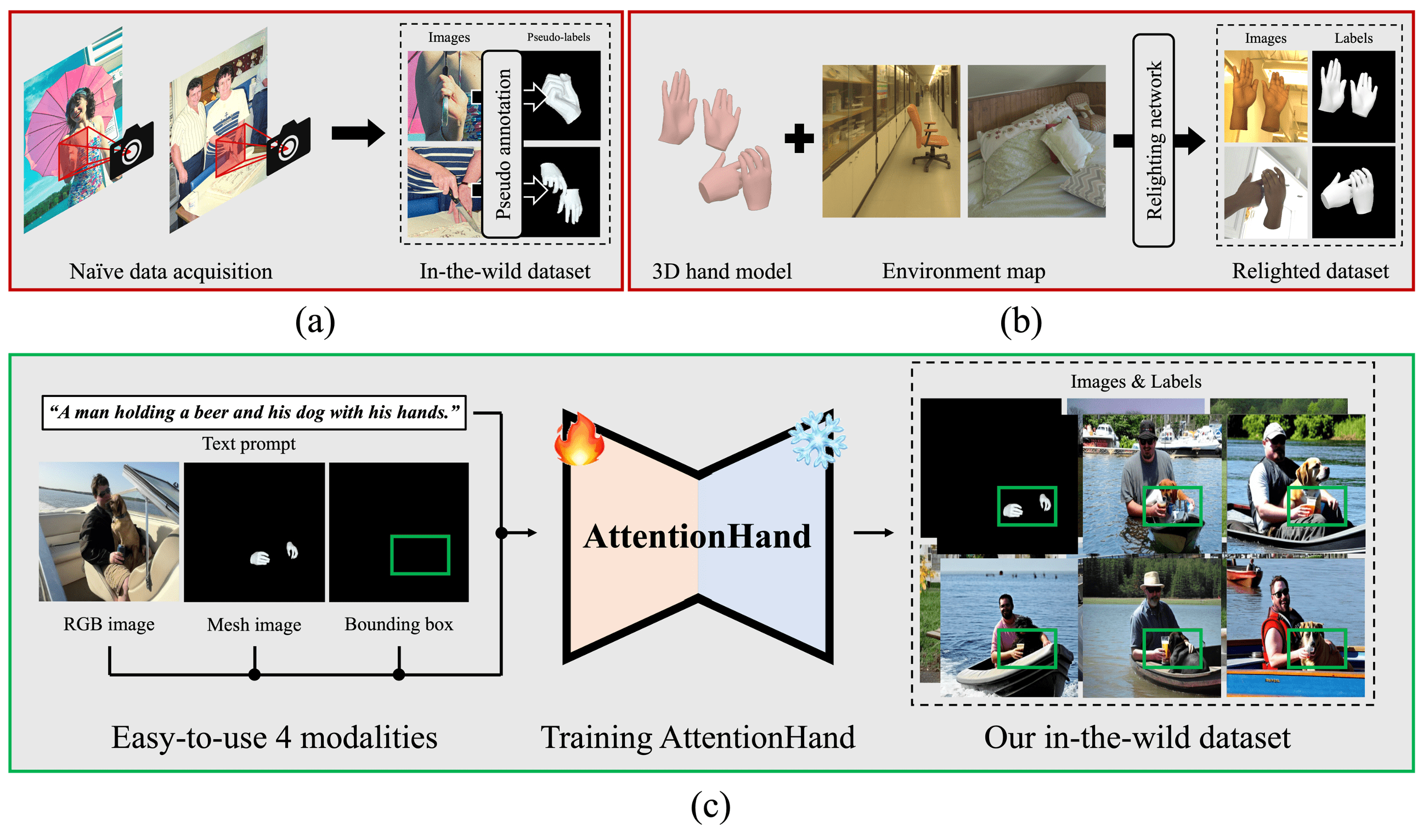}
\end{center}
\caption{Various acquisition types of 3D hand datasets.
(a) In-the-wild dataset (i.e., MSCOCO \cite{lin2014microsoft}) is naively acquired with inaccurate pseudo annotation, 
(b) relighted dataset (i.e., Re:InterHand \cite{moon2023dataset}) consists of unnatural hands with inharmonious background, and
(c) our in-the-wild dataset from AttentionHand, which is annotated with accurate 3D labels, contains natural hands with harmonious background, easy to generate, and can be made infinitely.}
\label{intro}
\end{figure*}

The goal of 3D hand mesh reconstruction is to recover the 3D hand mesh from a single RGB image. It becomes difficult when hands are in the wild, due to insufficiency of in-the-wild 3D hand datasets. 
Compared to in-the-lab datasets \cite{hampali2020honnotate, chao2021dexycb, ohkawa2023assemblyhands}, acquisition in-the-wild datasets is challenging due to unpredictable conditions such as weather, lighting, cost of sensors, and safety issues on crowded roads and public places. Even if an in-the-wild dataset is collected, data diversity would be poor due to the aforementioned severe constraints. Although arbitrary labels can be obtained through pseudo annotation, the precision and accuracy is still poor compared to in-the-lab datasets as shown in Fig. \ref{intro}(a).
To tackle this problem, several synthetic datasets \cite{lin2021two, moon2023dataset} have introduced. However, since the hand and background images are synthesized out of harmony, they consist of unnatural and unrealistic hand images as shown in Fig. \ref{intro}(b). Hence, it is difficult to overcome the domain gap between indoor and outdoor scenes with synthetic datasets.

Moreover, when hands are in a complex pose like interacting hands, it becomes even more challenging to reconstruct 3D hand meshes due to the appearance similarity, self-handed occclusion and depth ambiguity. 
Starting with InterHand2.6M \cite{moon2020interhand2}, several works \cite{rong2021monocular, zhang2021interacting, li2022interacting, hampali2022keypoint, meng20223d, ren2023decoupled, zuo2023reconstructing, li2023renderih} have emerged to solve the complex hand pose. However, they have been employed and evaluated primarily on in-the-lab scenes except for InterWild \cite{moon2023bringing}; it tried to relieve the domain gap by leveraging the geometric features of the hand, which is not affected by the domain. Nevertheless, since InterWild was trained with MSCOCO \cite{lin2014microsoft}, which is extremely lack of in-the-wild hand images with inaccurate 3D labels, there is earnestly need of more diverse, numerous, and accurately annotated in-the-wild datasets.

\begin{figure*}[t!]
\centering
\begin{center}
\includegraphics[width=\linewidth]{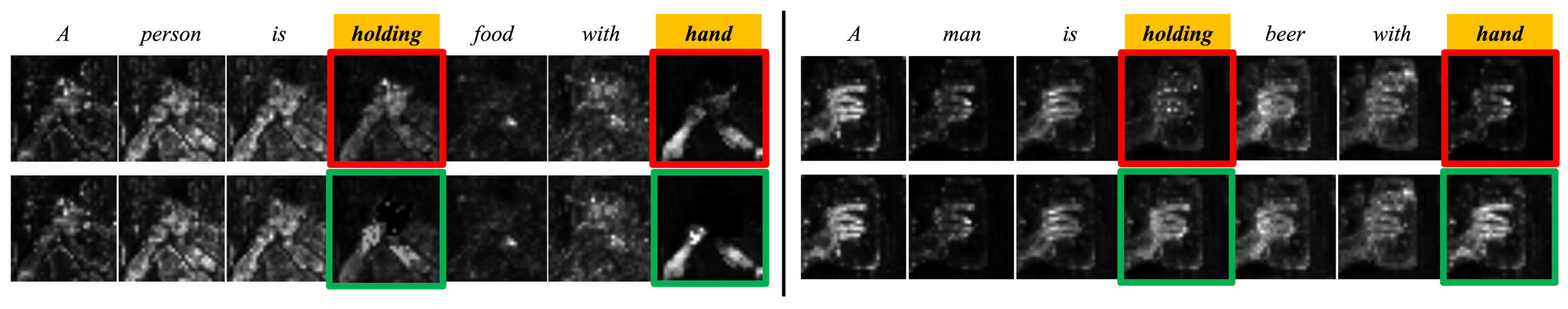}
\end{center}
\caption{Visualization of attention maps with corresponding tokens from given text prompts. Red and green boxes represent attention maps without and with AttentionHand, respectively.}
\label{attention}
\end{figure*}

To address aforementioned issues, we propose AttentionHand, a new method for the text-driven controllable hand image generation. AttentionHand is designed based on Stable Diffusion (SD) \cite{rombach2022high} to create accurate, natural, realistic and harmonious in-the-wild hand images easily and infinitely as shown in Fig. \ref{intro}(c).
AttentionHand has a huge advantage: we can simply generate images with only four modalities -- an RGB image, the corresponding hand mesh image, bounding box, and text prompt. 
Therefore, we can generate (1) various in-the-wild hand images with flexible text prompts, and (2) well-aligned hand images with 3D hand label. 
By generating new samples with AttentionHand, we can alleviate aforementioned issues of the 3D hand mesh reconstruction in the wild. 

To train AttentionHand, we need to additionally prepare a local RGB image and local mesh image for attention of hand-focused region of the image. The preparation of local information is essential because hands commonly occupy relatively small region in the image. Hence, we obtain local RGB and mesh images by cropping and resizing original RGB and mesh images (i.e., we define them as global information) with the bounding box of hand region. After encoding prepared information in the encoding phase, encoded latent embeddings are fed to the conditioning phase, which is composed of the text attention stage (TAS) and visual attention stage (VAS). 

TAS attends on hand-related tokens from the given text prompt by leveraging attention maps as shown in Fig. \ref{attention}. Specifically, TAS extracts hand-related attention maps (i.e., \textit{holding} and \textit{hand}), and these attention maps are updated to highlight hand-related regions by the refinement based on the softmax operation and Gaussian filter. With TAS, we can obtain more hand-focused images than before.
On the other hand, VAS attends on hand-related regions by conditioning global and local hand mesh images with the SD-based pipeline. With global and local information, AttentionHand can be jointly optimized to reflect the global context (i.e., in-the-wild background) and local context (i.e., hand-focused foreground.)
In the end of the conditioning phase, we finally get the diffusion feature, which is decoded to new hand images in the decoding phase. 
Hence, AttentionHand can generate well-aligned hand images with the given mesh image and text prompt for the 3D hand mesh reconstruction in the wild. 

To prove the excellence of AttentionHand, we conducted extensive experiments for the text-to-hand image generation and 3D hand mesh reconstruction. As a result, AttentionHand achieved state-of-the-art in the text-to-hand image generation, and the performance of 3D hand mesh reconstruction was considerably improved by additionally training with new hand samples generated by AttentionHand. Especially, the performance was enhanced significantly on in-the-wild datasets, which implies AttentionHand can generate various and well-annotated in-the-wild hand images. The summary of our contributions is as follows:
\begin{itemize}
    \item We propose a novel method, AttentionHand, which generates well-aligned in-the-wild hand images in a simple manner without laborious data acquisition.
    \item AttentionHand is designed based on a generative model that attends on hand-related tokens from the text prompt and hand-related regions from the hand mesh image, for generating hand-focused images.
    \item AttentionHand achieved state-of-the-art in the text-to-hand image generation, and we verified that utilizing the dataset generated from AttentionHand improves the performance on 3D hand mesh reconstruction in the wild.
\end{itemize}

\section{Related Work}

\subsection{Text-to-Image Generation}
Text-to-image generation aims to synthesize high-resolution image from natural language descriptions.
With the advent of diffusion models, various studies on text-to-image generation have been conducted in recent years \cite{rombach2022high, zhang2023adding, mou2023t2i, zhao2023uni, podell2023sdxl}.
Specifically, ControlNet \cite{zhang2023adding} and T2I-Adapter \cite{mou2023t2i} proposed novel approaches to incorporate arbitrary condition into the generation process.
Recently, Uni-ControlNet \cite{zhao2023uni} presented a novel approach that allows for the simultaneous utilization of various conditions in a flexible and composable manner.
Nevertheless, aforementioned models exhibited common limitations in generating hand images, due to the relatively small size of hands within the overall image resolution.

\subsection{Generative Models for Hand}

\subsubsection{GANs for Hand.}
There are several works \cite{mueller2018ganerated, tang2018gesturegan, hu2021model, hu2022hand} to tackle the hand image generation problem with the generative adversarial network (GAN) \cite{goodfellow2014generative}. 
Specifically, a novel network for image-to-image translation \cite{mueller2018ganerated} was proposed to make generated images follow the same statistical distribution as real-world hand images.
GestureGAN \cite{tang2018gesturegan} was designed to translate hand gesture-to-gesture with the explicit hand skeleton information through the color loss and the cycle-consistency loss.
Moreover, the first model-aware gesture-to-gesture translation framework \cite{hu2021model} was introduced with hand prior as the intermediate representation.
Recently, a new method \cite{hu2022hand}, which employs the expressive model-aware hand-object representation and leverages its inherent topology to build the unified surface space, was proposed.
However, these works have a common limitation; they are confined to target gestures in generating new hand images.
In other words, they are inappropriate to generate in-the-wild images focused on various hands.

\subsubsection{Diffusion Models for Hand.}
Recently, some works \cite{li2023diffhand, lin2023handdiffuse, lu2023handrefiner} have been addressed hand-related problems with diffusion models.
DiffHand \cite{li2023diffhand} introduced the first diffusion-based framework that approaches hand mesh reconstruction as a denoising diffusion process. 
HandDiffuse \cite{lin2023handdiffuse} proposed a strong baseline for the controllable motion generation of interacting hands using various controllers by designing a diffusion-based model.
HandRefiner \cite{lu2023handrefiner} presented an inpainting pipeline to rectify malformed human hands in generated images with diffusion-based models. 
However, since these models are not text-driven methods, they cannot generate various in-the-wild hand images conditioned on language instructions. 

\section{Method}

We introduce AttentionHand, a novel framework for creating various and plausible hand images. AttentionHand is a SD-based framework that can generate new RGB images infinitely conditioned on hand mesh images and text prompts. The overall pipeline is shown in Fig. \ref{pipeline}.

\subsection{Data Preparation Phase}
As shown in the first box of Fig. \ref{pipeline}, it just requires four inputs to train AttentionHand: (1) a global RGB hand image $I_{RGB}^G\in \mathbb{R}^{3\times512\times512}$, (2) a global hand mesh image $I^{G}_{mesh} \in \mathbb{R}^{3 \times 512 \times 512}$, (3) a bounding box of the hand region $B \in \mathbb{R}^{1 \times 4}$, and (4) a hand-related text prompt $U$. However, since hands typically occupy small areas on in-the-wild scenes, we also obtain a local RGB hand image $I_{RGB}^L \in \mathbb{R}^{3 \times 512 \times 512}$ and a local hand mesh image $I^{L}_{mesh} \in \mathbb{R}^{3 \times 512 \times 512}$ by cropping and resizing $I_{RGB}^G$ and $I^{G}_{mesh}$ with $B$. This combination of local and global information enhances hand image conditioning. Details will be explained in the supplementary materials.

\subsection{Encoding Phase}
\begin{figure*}[t!]
\centering
\begin{center}
\includegraphics[width=\linewidth]{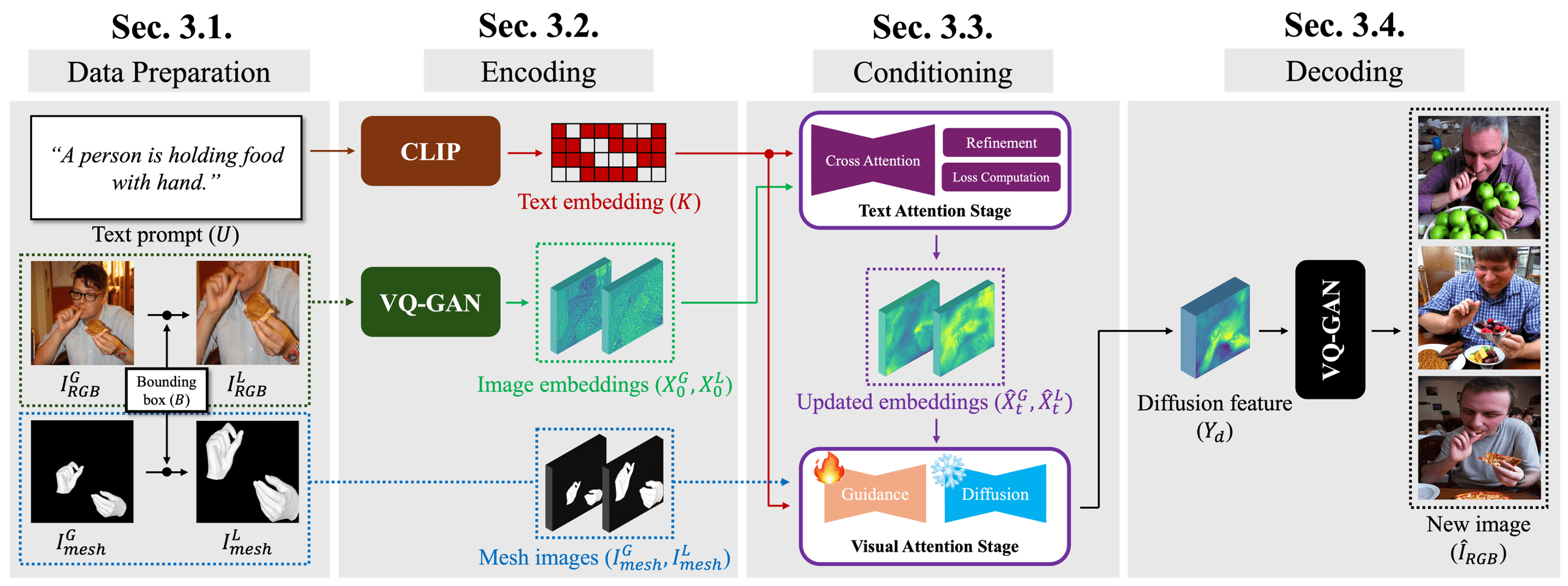}
\end{center}
\caption{Overall pipeline of AttentionHand. 
In the data preparation phase, we prepare global and local RGB images, global and local hand mesh images, bounding box, and text prompt.
In the encoding phase, we get global and local latent image embeddings through VQ-GAN \cite{esser2021taming}, and text embedding through CLIP \cite{radford2021learning}. 
In the conditioning phase, we refine image embeddings through the text attention stage, and obtain the diffusion feature through the visual attention stage. 
In the decoding phase, we generate a new hand image $\hat{I}_{RGB}$ from $Y_d$ through VQ-GAN.
}
\label{pipeline}
\end{figure*}

For the diffusion process in latent space, encoding phase for $I_{RGB}^G$, $I_{RGB}^L$ and $U$ is implemented by the encoder $\mathcal{E}$. It makes global and local latent image embeddings $X_0^G,X_0^L \in \mathbb{R}^{4\times64\times64}$ for $I_{RGB}^G$ and $I_{RGB}^L$, and a latent text embedding $K \in \mathbb{R}^{77\times768}$ for $U$. Specifically, $X_0^G$ and $X_0^L$ are obtained by VQ-GAN \cite{esser2021taming}, and $K$ is obtained by CLIP \cite{radford2021learning} as shown in the second box of Fig. \ref{pipeline}. These latent embeddings are fed as inputs to the conditioning phase, which will be introduced by the next subsection. The encoding phase is expressed as follows:
\begin{equation}
X_0^G,X_0^L,K=\mathcal{E}(I_{RGB}^G,I_{RGB}^L,U).
\end{equation}

\subsection{Conditioning Phase}
For generating new hand images conditioned by given text prompt and mesh images, we design the text attention stage (TAS) and the visual attention stage (VAS) in the conditioning phase, as shown in the third box of Fig. \ref{pipeline}. TAS is a stage of paying attention to tokens for the hand and its corresponding gesture in a given text. VAS is a stage of training the SD-based model specialized for hand image generation by conditioning global and local mesh images.

\subsubsection{Text Attention Stage (TAS).}

\begin{figure*}[t!]
\centering
\begin{center}
\includegraphics[width=\linewidth]{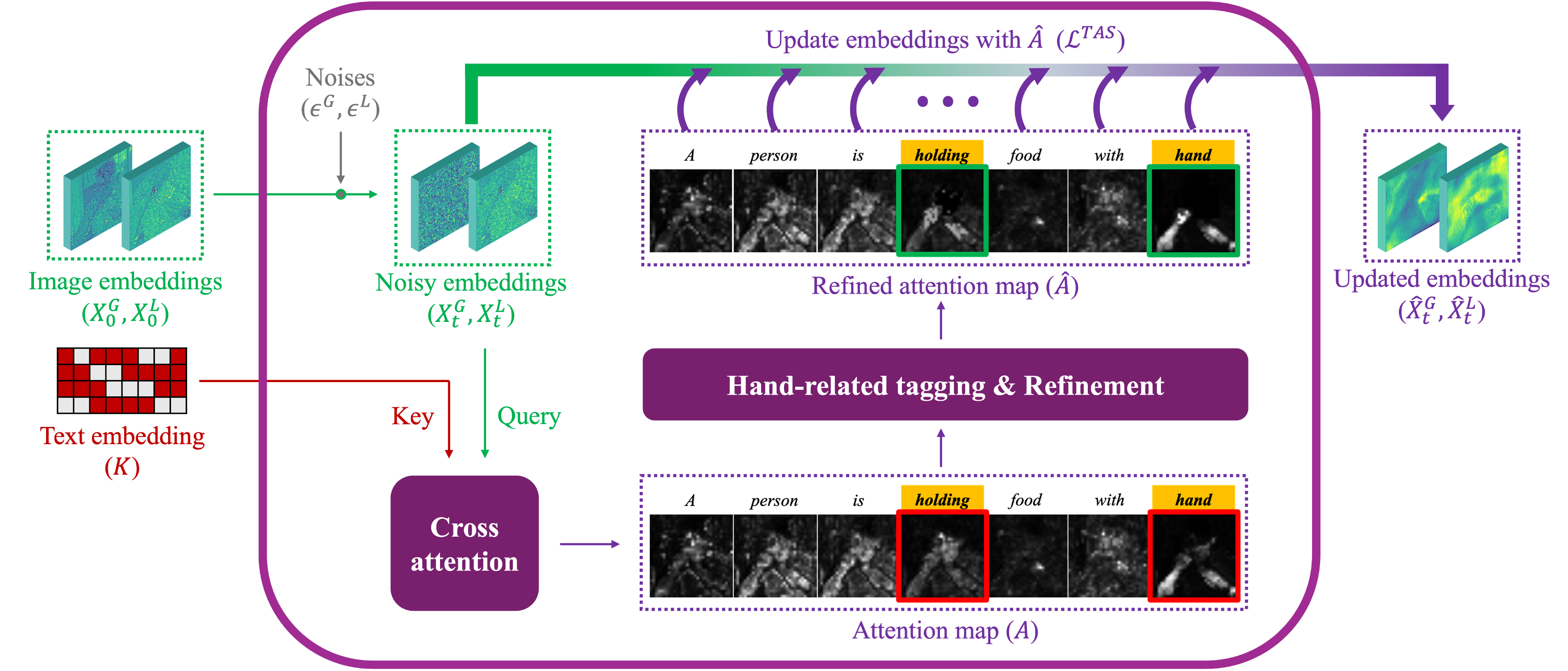}
\end{center}
\caption{Overall process of the text attention stage (TAS). By leveraging the hand-related tagging and refinement, we can highlight hand-related attention maps, which leads to update noisy embeddings with $\mathcal{L}^{TAS}$.}
\label{TAS}
\end{figure*}

TAS is a stage of attending tokens which represent hand or gestures in a given text prompt as shown in Fig. \ref{TAS}.
First, by adding Gaussian noise to $X_0^{G}$ and $X_0^{L}$ with $t$ diffusion steps, the global noisy embedding $X_t^{G}$ and local noisy embedding $X_t^{L}$ are obtained. For simplicity, we define as $X_0 = (X_0^{G}, X_0^{L})$ and $X_{t} = (X_t^{G},X_t^{L})$.
Then, $X_t$ and $K$ are fed to TAS as inputs. 
For the text attention of TAS, we utilize the cross attention \cite{chen2021crossvit}. Specifically, an attention map $A \in \mathbb{R}^{H \times W \times N}$ is obtained by calculating the key (i.e., $K$) and query (i.e., $Q$, which is the linear projection of intermediate feature map from $X_{t}$ of U-Net \cite{ronneberger2015u} in SD). $H$ and $W$ denote the height and width of $A$, and $N$ denotes the number of all tokens of $K$.

Next, to extract hand-related attention maps $A_{k \in K} \in \mathbb{R}^{H \times W \times {N_k}}$ in $A$, we design the hand-related tagging $\mathcal{H}_{tag}$, which is based on part-of-speech tagging \cite{chiche2022part}, where $N_k$ denotes the number of hand-related tokens in $K$. Specifically, $\mathcal{H}_{tag}$ determines if the input token indicates the hand-related word (i.e., \textit{holding}, \textit{taking}, or \textit{hand}). With $\mathcal{H}_{tag}$, we can attend hand-related tokens $k$ to generate more hand-focused images. More details are in the supplementary materials.

Then, we employ the softmax operation and Gaussian smoothing to maximize the effect of $A_k$. Since the Gaussian filter effectively removes noise from images and preserves detailed information by using the average value of surrounding pixels, we fully exploit these advantages. Hence, $A_k$ is updated to $\hat{A}_k \in \mathbb{R}^{H \times W \times {N_k}}$ by refining hand-related attention maps as follows:
\begin{equation}
    \hat{A}_k = Gaussian(Softmax(A_k)).
\end{equation}
For simplicity, we define $\hat{A} \in \mathbb{R}^{H \times W \times N}$ as the concatenation of $\hat{A}_{k \in K}$ and $A_{l \notin K} \in \mathbb{R}^{H \times W \times {N_l}}$, where $N_l$ denotes the number of not hand-related tokens in $K$, and $N={N_k}+{N_l}$.

Moreover, optimization for evenly reflecting the image features of all attention maps is necessary. In other words, it is required to design an objective to prevent poor generation of the image feature for a specific token. 
Specifically, for arbitrary token $n \in K$, the highest value $s_n$ among all patches in the $n$-th refined attention map $\hat{A}_n$ is extracted, and it is subtracted from 1. This operation is implemented for all tokens in $K$, and a novel loss, which named $\mathcal{L}^{TAS}$, is computed the largest value among them as follows:
\begin{equation}
\mathcal{L}^{TAS} = max_{n \in K}(1-s_n).
\end{equation}
Based on $\mathcal{L}^{TAS}$, $X_t$ is updated to $\hat{X}_t = (\hat{X}_t^G, \hat{X}_t^L)$ as follows:
\begin{equation}\label{eqn:TAS}
\hat{X}_t = {X}_t - \alpha_t \nabla_{{X}_t}\mathcal{L}^{TAS},
\end{equation}
where $\alpha_t$ indicates the learning rate, and $\hat{X}_t^G$ and $\hat{X}_t^L$ indicate updated global and local noisy embedding, respectively.

\begin{figure*}[t!]
\centering
\begin{center}
\includegraphics[width=0.6\linewidth]{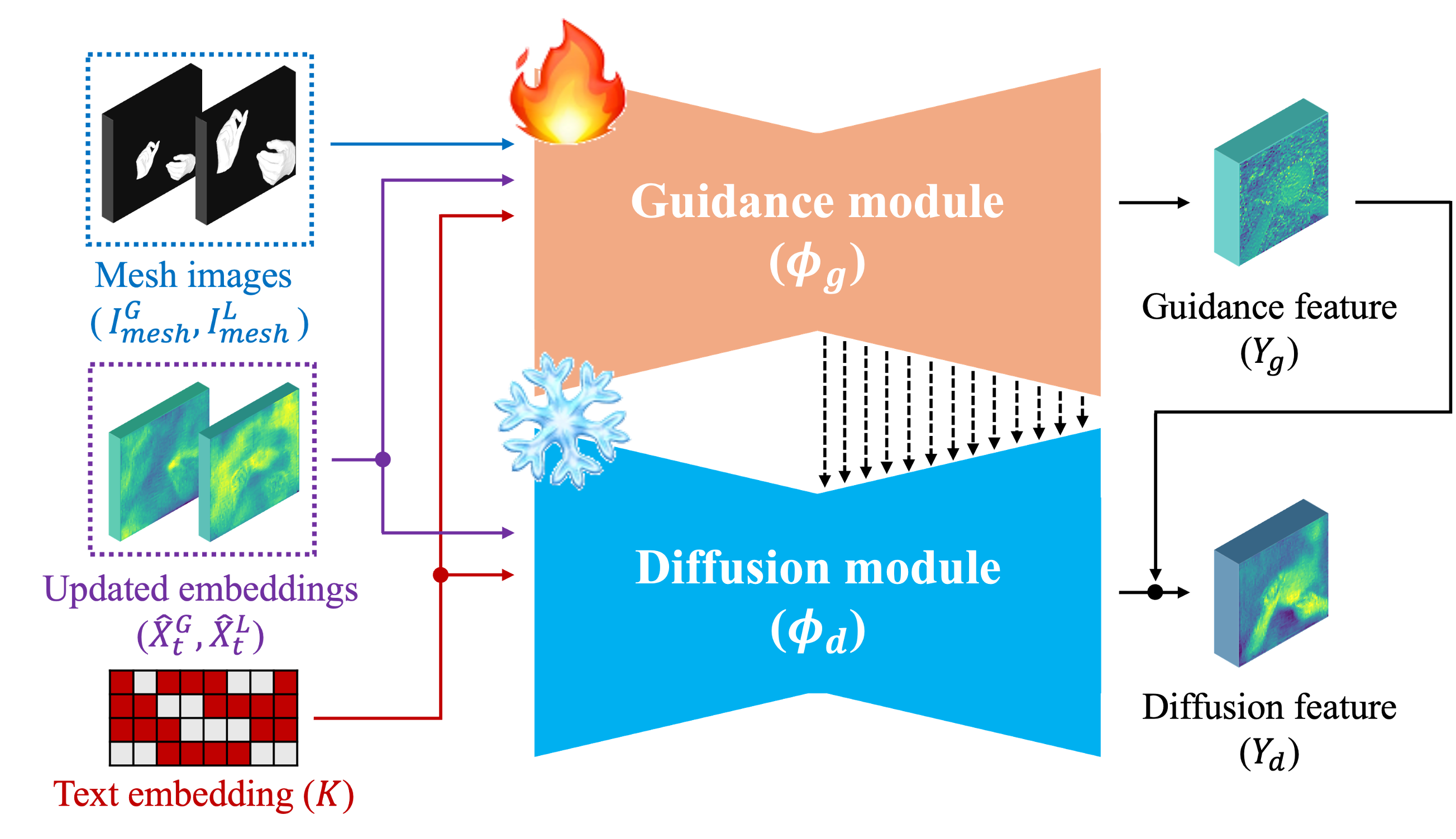}
\end{center}
\caption{Overall process of the visual attention stage (VAS). By utilizing the global and local information, we can obtain the harmonious diffusion feature, which leads to generate high-fidelity hand images.}
\label{VAS}
\end{figure*}

\subsubsection{Visual Attention Stage (VAS).}

VAS is a stage of training SD-based model by conditioning the aforementioned global and local mesh image. VAS is composed of two modules: one is the guidance module $\phi_{g}$, and the other is the diffusion module $\phi_{d}$ as shown in Fig. \ref{VAS}. 
First, the diffusion module $\phi_{d}$ is designed based on an U-Net network, consisting of 25 blocks: 8 blocks are downsampling and upsampling convolution layers, and the remaining 17 blocks consist of four ResNet \cite{he2016deep} layers and two Vision Transformers \cite{dosovitskiy2020image}. We define the parameter set of $\phi_{d}$ as $\theta_d$, which is fixed frozen to maintain the image generation performance of SD. 

On the other hand, the guidance module $\phi_g$ is also based on an U-Net network with 25 blocks of $\phi_d$. We define the parameter set of $\phi_g$ as $\theta_g$, which is a copied version of $\theta_d$. Different from $\theta_d$, $\theta_g$ is set to be learnable for generating images conditioned to $I^{G}_{mesh}$ and $I^{L}_{mesh}$. Specifically, $\phi_g$ has zero convolution $\mathcal{Z}$ \cite{zhang2023adding} at the front of the network, and last 12 blocks of the network consist of $\mathcal{Z}$. Since $\mathcal{Z}$ is defined as a $1 \times 1$ convolution layer whose weights and bias are initialized to zero, the gradients of the weight and bias progressively grow from zeros to optimized parameters in a learnable manner. Hence, $\mathcal{Z}$ helps generated images to be conditioned on $I^{G}_{mesh}$ and $I^{L}_{mesh}$, while maintaining the quality of image generation.
More specifically, $\phi_d$ and $\phi_g$ share weights at the beginning of training, because parameter sets of both modules, i.e, $\theta_d$ and $\theta_g$, are initialized with the pre-trained SD. However, while continuing with training process, $\theta_g$ is updated to learn $I^{G}_{mesh}$ and $I^{L}_{mesh}$, whereas $\theta_d$ is fixed frozen to preserve the performance of image generation. At the end of training, $\theta_d$ and $\theta_g$ are completely different from the beginning. Hence, $\phi_g$ is formulated as follows:
\begin{equation}
Y_{g}=\phi_g(\hat{X}_t,I_{mesh},K,t;\theta_g),
\end{equation}
where $I_{mesh}=(I^{G}_{mesh},I^{L}_{mesh})$ denotes the concatenation of the global and local mesh image, $t$ denotes the diffusion step obtained by positional encoding, $Y_{g}=(Y_{g}^G,Y_{g}^L) \in \mathbb{R}^{2\times4\times64\times64}$ denotes the concatenation of the global guidance feature $Y_{g}^G$ and local guidance feature $Y_{g}^L$. 
Next, $\phi_d$ is formulated as follows:
\begin{equation}
Y_{d}=\phi_{d}(\hat{X}_t,K,t;\theta_d)+Y_{g},
\end{equation}
where $Y_{d}=(Y_{d}^G,Y_{d}^L)\in \mathbb{R}^{2\times4\times64\times64}$ indicates the concatenation of the global diffusion feature $Y_{d}^G$ and local diffusion feature $Y_{d}^L$.

\begin{figure*}[t!]
\centering
\begin{center}
\includegraphics[width=0.6\linewidth]{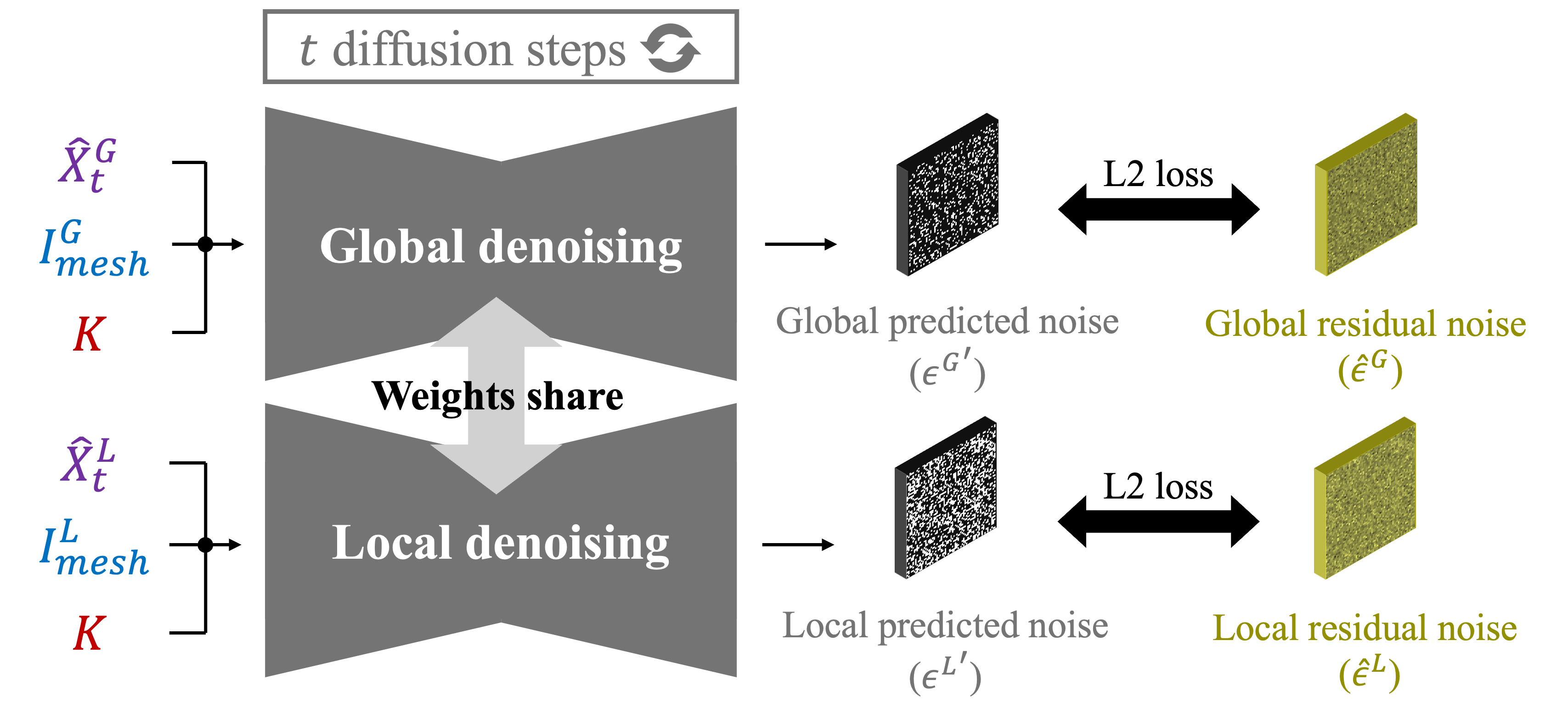}
\end{center}
\caption{Overall process of the optimization of AttentionHand. By global and local denoising of updated noisy embeddings with $t$ diffusion steps, we obtain global and local predicted noises. They are optimized by L2 loss with global and local residual noises. Note that global and local denoising networks share weights.}
\label{optimization}
\end{figure*}

\subsubsection{Optimization.}

Since the diffusion model typically involves both forward and reverse processes, our AttentionHand also employs two processes. 
For the forward process, the noisy embedding $X_t = (X_t^{G},X_t^{L})$ is obtained by progressively perturbing Gaussian noise $\epsilon = (\epsilon^G, \epsilon^L)$ to the initial embedding $X_0 = (X_0^G, X_0^L)$ by $t$ diffusion steps. $\epsilon^G$ and $\epsilon^L$ denote the global and local noise added to $X_0^G$ and $X_0^L$, respectively. Then, since $X_t$ is updated to $\hat{X}_t$ by TAS, $\epsilon$ is also updated to $\hat{\epsilon} = (\hat{\epsilon^G}, \hat{\epsilon^L})$. In other words, $\hat{\epsilon}$ is considered the residual noise between $X_0$ and $\hat{X}_t$. Thus, $\hat{\epsilon^G}$ and $\hat{\epsilon^L}$ denote the global and local residual noise.
For the reverse process, AttentionHand learns to gradually remove residual noises with global and local denoising processes as shown in Fig. \ref{optimization}. Therefore, given text embedding $K$, diffusion steps $t$, and mesh images $I^{G}_{mesh}$ and $I^{L}_{mesh}$, the diffusion training network $\epsilon_\theta$ is optimized to predict $\hat{\epsilon^G}$ and $\hat{\epsilon^L}$ jointly through the following objectives:
\begin{equation}
\mathcal{L}^{G}=\mathbb{E}_{X^{G}_0,I^G_{mesh},K,t,\hat{\epsilon^{G}} \sim \mathcal{N}(0,1)}[\|\hat{\epsilon^{G}}-\epsilon_{\theta}(\hat{X}^{G}_{t},I^G_{mesh},K,t)\|^2_2],
\end{equation}
\begin{equation}
\mathcal{L}^{L}=\mathbb{E}_{X^{L}_0,I^L_{mesh},K,t,\hat{\epsilon^{L}} \sim \mathcal{N}(0,1)}[\|\hat{\epsilon^{L}}-\epsilon_{\theta}(\hat{X}^{L}_{t} ,I^L_{mesh},K,t)\|^2_2],
\end{equation}
where $\mathcal{L}^{G}$ and $\mathcal{L}^{L}$ indicate the cost function of global and local features, respectively. Thus, the final objective is defined as follows:
\begin{equation}
\mathcal{L}=\lambda^{G} \mathcal{L}^{G}+\lambda^{L} \mathcal{L}^{L},
\end{equation}
where $\lambda^{G}$ and $\lambda^{L}$ are weighted coefficients of $\mathcal{L}^{G}$ and $\mathcal{L}^{L}$.

\subsection{Decoding Phase}
In the decoding phase, we can generate a new RGB hand image $\hat{I}_{RGB} \in \mathbb{R}^{3 \times 512 \times 512}$ by passing $Y_d$ through the decoder $\mathcal{D}$, as shown in the fourth box of Fig. \ref{pipeline}. While $\mathcal{E}$ encodes $X_0$ by downsampling $I_{RGB}$ in the latent space, $\mathcal{D}$ decodes $\hat{I}_{RGB}$ by upsampling $Y_d$ in the pixel space, conditioned to given text prompt and mesh images. The structure of $\mathcal{D}$ is similar to the decoder of VQ-GAN. The decoding phase is expressed as follows:
\begin{equation}
\hat{I}_{RGB}=\mathcal{D}(Y_{d}).
\end{equation}

\section{Experiments}

\subsection{Datasets}
For the text-to-image generation, we adopted MSCOCO \cite{lin2014microsoft}.
For the 3D hand mesh reconstruction, we adopted Hands-In-Action (HIC) \cite{tzionas2016capturing}, Re:InterHand (ReIH) \cite{moon2023dataset}, InterHand2.6M (IH2.6M) \cite{moon2020interhand2}, and MSCOCO. 
Due to the page limit, details will be explained in the supplementary materials.

\subsection{Evaluation Protocol}
For the text-to-image generation, we adopted FID \cite{heusel2017gans}, FID-Hand (FID-H), KID \cite{binkowski2018demystifying}, KID-Hand (KID-H), the hand confidence score (Hand Conf.) \cite{narasimhaswamy2024handiffuser}, the mean square error of 2D and 3D keypoints (MSE-2D, 3D), and the user preference (User Pref.). 
For the 3D hand mesh reconstruction, we adopted the mean per-vertex position error (MPVPE), the right hand-relative vertex error (RRVE), and the mean relative-root position error (MRRPE).
Due to the page limit, details will be explained in the supplementary materials.

\begin{table}[t!]
\caption{Quantitative comparisons with state-of-the-art text-to-image generation models.}
\label{tab:gen}
\centering
\resizebox{\linewidth}{!}{
\begin{tabular}{l|cccccccc}
\toprule
{Methods} & {FID$\downarrow$} & {KID$\downarrow$} & \textcolor{black}{FID-H$\downarrow$} & \textcolor{black}{KID-H$\downarrow$} & \textcolor{black}{Hand Conf.$\uparrow$} & \textcolor{black}{MSE-2D$\downarrow$} & {MSE-3D$\downarrow$} & {User Pref.(\%)$\uparrow$} \\
\hline
\hline
{Stable Diffusion \cite{rombach2022high}} & {40.52} & {0.00684} & \textcolor{black}{50.78} & \textcolor{black}{0.02554} & \textcolor{black}{0.651} & \textcolor{black}{2.932} & {4.591} & {5.864} \\
{Uni-ControlNet \cite{zhao2023uni}} & {30.34} & {0.00744} & \textcolor{black}{37.77} & \textcolor{black}{0.02004} & \textcolor{black}{0.855} & \textcolor{black}{2.105} & {3.039} & {8.796} \\
{T2I-Adapter \cite{mou2023t2i}} & {22.00} & {0.00761} & \textcolor{black}{32.08} & \textcolor{black}{0.01568} & \textcolor{black}{0.914} & \textcolor{black}{1.546} & {2.451} & {19.676} \\
{ControlNet \cite{zhang2023adding}} & {21.67} & {0.00658} & \textcolor{black}{40.32} & \textcolor{black}{0.02098} & \textcolor{black}{0.810} & \textcolor{black}{1.252} & {2.182} & {7.948} \\
{AttentionHand (w/o TAS)} & {21.27} & {0.00331} & \textcolor{black}{28.56} & \textcolor{black}{0.01390} & \textcolor{black}{0.955} & \textcolor{black}{1.211} & {2.042} & {20.734} \\
\rowcolor{Gray}
\textbf{AttentionHand (w/ TAS)} & \textbf{20.71} & \textbf{0.00301} & \textbf{\textcolor{black}{27.09}} & \textbf{\textcolor{black}{0.01287}} & \textbf{\textcolor{black}{0.965}} & \textbf{\textcolor{black}{1.026}} & \textbf{1.986} & \textbf{36.905} \\
\bottomrule
\end{tabular}
}
\end{table}

\begin{figure*}[t!]
\centering
\begin{center}
\includegraphics[width=\linewidth]{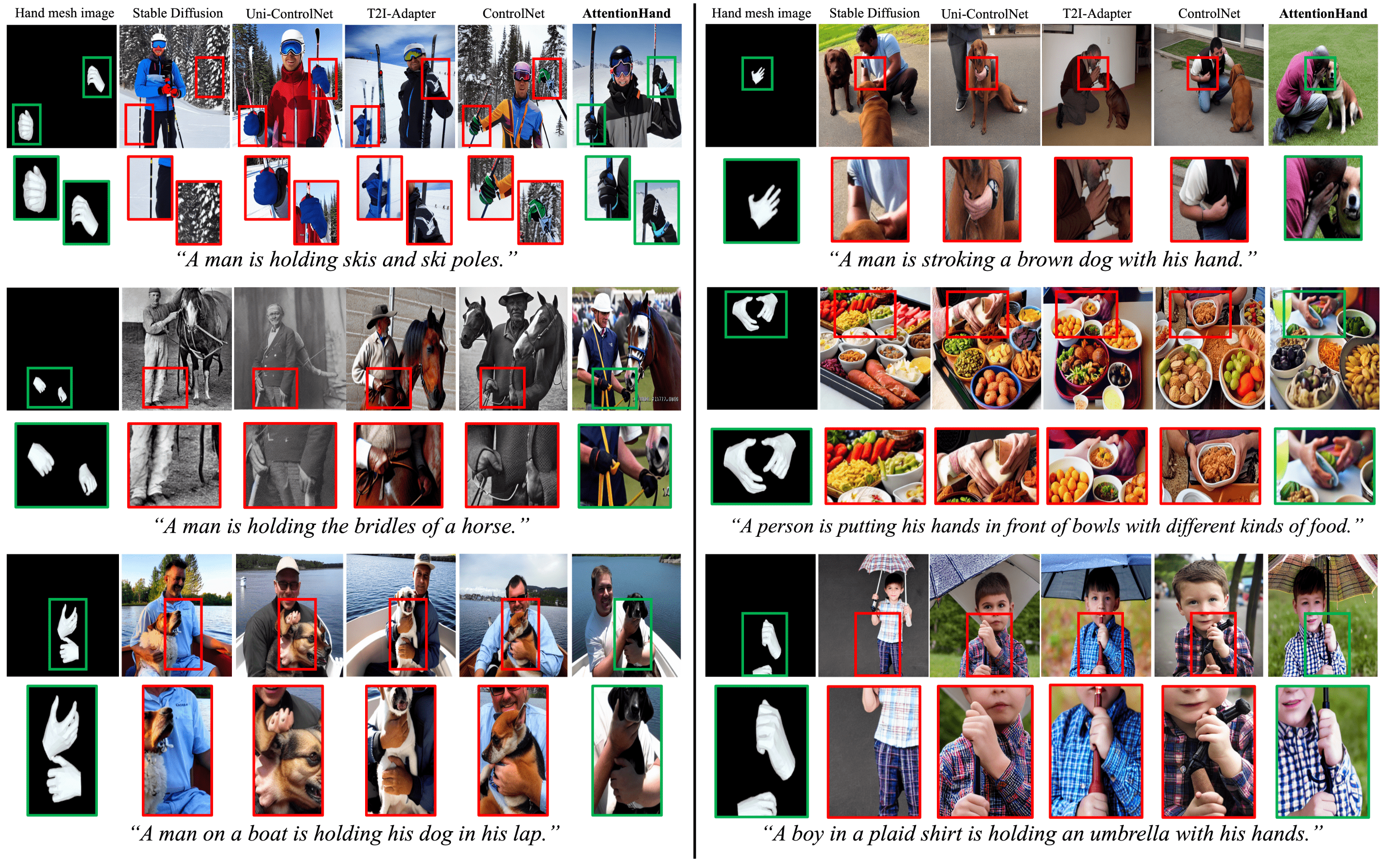}
\end{center}
\caption{Qualitative comparisons with state-of-the-art text-to-image generation models. 
Red and green boxes in each sample indicate the wrong and corrent hand bounding box, respectively. 
}
\label{exp_gen}
\end{figure*}

\subsection{Comparisons with State-of-the-arts}

\subsubsection{Text-to-Image Generation.}
As shown in Table \ref{tab:gen}, our AttentionHand exhibited the highest performance in all metrics among state-of-the-arts \cite{rombach2022high, zhao2023uni, mou2023t2i, zhang2023adding}. This is particularly evident in the comparison of FID(-H) and KID(-H), which signify the quality of the generated images being on par with real RGB images. Furthermore, the lowest MSE-2D and MSE-3D indicates the remarkable alignment between the generated images and the corresponding hand mesh images. With respect to the user preference, AttentionHand scored the highest compared to other methods. It implies that most users acknowledged the outstanding quality of hand images generated by AttentionHand. 
In addition, as shown in Fig. \ref{exp_gen}, our AttentionHand generated the high-quality hand image which is well-corresponding with the given mesh image and fully reflected the given text prompt. Specifically, even when two-hands mesh image is given, which is more challenging than in the case of single-hand mesh image, AttentionHand generated the hand image robustly. It implies our AttentionHand is proper to generate well-aligned hand images with given mesh images and text prompt. Additional qualitative results are in the supplementary materials.

\begin{table}[t!]
\caption{Quantitative comparisons with state-of-the-art 3D hand mesh reconstruction methods with and without AttentionHand. The red subscripts indicate the difference in performance with and without AttentionHand.}
\label{tab:mesh}
\centering
\resizebox{\linewidth}{!}{
\begin{tabular}{l|ccc|ccc|ccc}
\toprule
\multicolumn{1}{r|}{Datasets} & \multicolumn{6}{c|}{In-the-wild} & \multicolumn{3}{c}{In-the-lab} \\
\cline{2-10}
{} & \multicolumn{3}{c|}{HIC \cite{tzionas2016capturing}} & \multicolumn{3}{c|}{ReIH \cite{moon2023dataset}} & \multicolumn{3}{c}{IH2.6M \cite{moon2020interhand2}} \\
\cline{2-10}
{Methods} & {MPVPE$\downarrow$} & {RRVE$\downarrow$} & {MRRPE$\downarrow$} & {MPVPE$\downarrow$} & {RRVE$\downarrow$} & {MRRPE$\downarrow$} &  {MPVPE$\downarrow$} & {RRVE$\downarrow$} & {MRRPE$\downarrow$}\\
\hline
\hline
{IHMR \cite{rong2021monocular}} & 38.57 & 45.51 & 119.64 & 30.90 & 45.55 & 98.45 & 16.94 & 21.98 & 33.39 \\
\rowcolor{Gray}
\textbf{IHMR+AttentionHand} & $\textbf{36.73}_{\textcolor{red}{-1.84}}$ & $\textbf{44.10}_{\textcolor{red}{-1.41}}$ & $\textbf{94.63}_{\textcolor{red}{-25.01}}$ & $\textbf{29.11}_{\textcolor{red}{-1.79}}$ & $\textbf{43.12}_{\textcolor{red}{-2.43}}$ & $\textbf{87.07}_{\textcolor{red}{-11.38}}$ & $\textbf{15.09}_{\textcolor{red}{-1.85}}$ & $\textbf{20.55}_{\textcolor{red}{-1.43}}$ & $\textbf{32.21}_{\textcolor{red}{-1.18}}$  \\
\hline
{InterShape \cite{zhang2021interacting}} & 27.66 & 34.69 & 110.25 & 27.87 & 38.56 & 80.04 & 12.97 & 17.35 & 31.56  \\
\rowcolor{Gray}
\textbf{InterShape+AttentionHand} & $\textbf{25.04}_{\textcolor{red}{-2.62}}$ & $\textbf{33.33}_{\textcolor{red}{-1.36}}$ & $\textbf{80.17}_{\textcolor{red}{-30.08}}$ & $\textbf{26.44}_{\textcolor{red}{-1.43}}$ & $\textbf{36.54}_{\textcolor{red}{-2.02}}$ & $\textbf{61.41}_{\textcolor{red}{-18.63}}$ & $\textbf{11.90}_{\textcolor{red}{-1.07}}$ & $\textbf{16.22}_{\textcolor{red}{-1.13}}$ & $\textbf{30.04}_{\textcolor{red}{-1.52}}$  \\
\hline
{IntagHand \cite{li2022interacting}} & 23.07 & 28.74 & 52.46 & 25.90 & 30.05 & 42.22 & 12.34 & 17.32 & 29.31  \\
\rowcolor{Gray}
\textbf{IntagHand+AttentionHand} & $\textbf{21.87}_{\textcolor{red}{-1.20}}$ & $\textbf{27.09}_{\textcolor{red}{-1.65}}$ & $\textbf{47.11}_{\textcolor{red}{-5.35}}$ & $\textbf{23.39}_{\textcolor{red}{-2.51}}$ & $\textbf{28.77}_{\textcolor{red}{-1.28}}$ & $\textbf{33.98}_{\textcolor{red}{-8.24}}$ & $\textbf{11.42}_{\textcolor{red}{-0.92}}$ & $\textbf{15.81}_{\textcolor{red}{-1.51}}$ & $\textbf{29.18}_{\textcolor{red}{-0.13}}$ \\
\hline
{DIR \cite{ren2023decoupled}} & 21.89 & 26.11 & 43.11 & 21.82 & 29.66 & 37.01 & 10.26 & 17.11 & 28.98  \\
\rowcolor{Gray}
\textbf{DIR+AttentionHand} & $\textbf{20.66}_{\textcolor{red}{-1.23}}$ & $\textbf{25.87}_{\textcolor{red}{-0.24}}$ & $\textbf{40.54}_{\textcolor{red}{-2.57}}$ & $\textbf{19.91}_{\textcolor{red}{-1.91}}$ & $\textbf{26.67}_{\textcolor{red}{-2.99}}$ & $\textbf{35.05}_{\textcolor{red}{-1.96}}$ & $\textbf{10.09}_{\textcolor{red}{-0.17}}$ & $\textbf{16.99}_{\textcolor{red}{-0.12}}$ & $\textbf{28.02}_{\textcolor{red}{-0.96}}$  \\
\hline
{InterWild \cite{moon2023bringing}} & 15.30 & 21.35 & 31.26 & 13.99 & 20.07 & 22.38 & 11.52 & 19.77 & 26.87  \\
\rowcolor{Gray}
\textbf{InterWild+AttentionHand} & $\textbf{14.74}_{\textcolor{red}{-0.56}}$ & $\textbf{21.10}_{\textcolor{red}{-0.25}}$ & $\textbf{29.26}_{\textcolor{red}{-2.00}}$ & $\textbf{13.95}_{\textcolor{red}{-0.04}}$ & $\textbf{19.94}_{\textcolor{red}{-0.13}}$ & $\textbf{22.05}_{\textcolor{red}{-0.33}}$ & $\textbf{10.62}_{\textcolor{red}{-0.90}}$ & $\textbf{19.09}_{\textcolor{red}{-0.68}}$ & $\textbf{25.74}_{\textcolor{red}{-1.13}}$  \\
\bottomrule
\end{tabular}
}
\end{table}

\begin{figure*}[t!]
\centering
\begin{center}
\includegraphics[width=\linewidth]{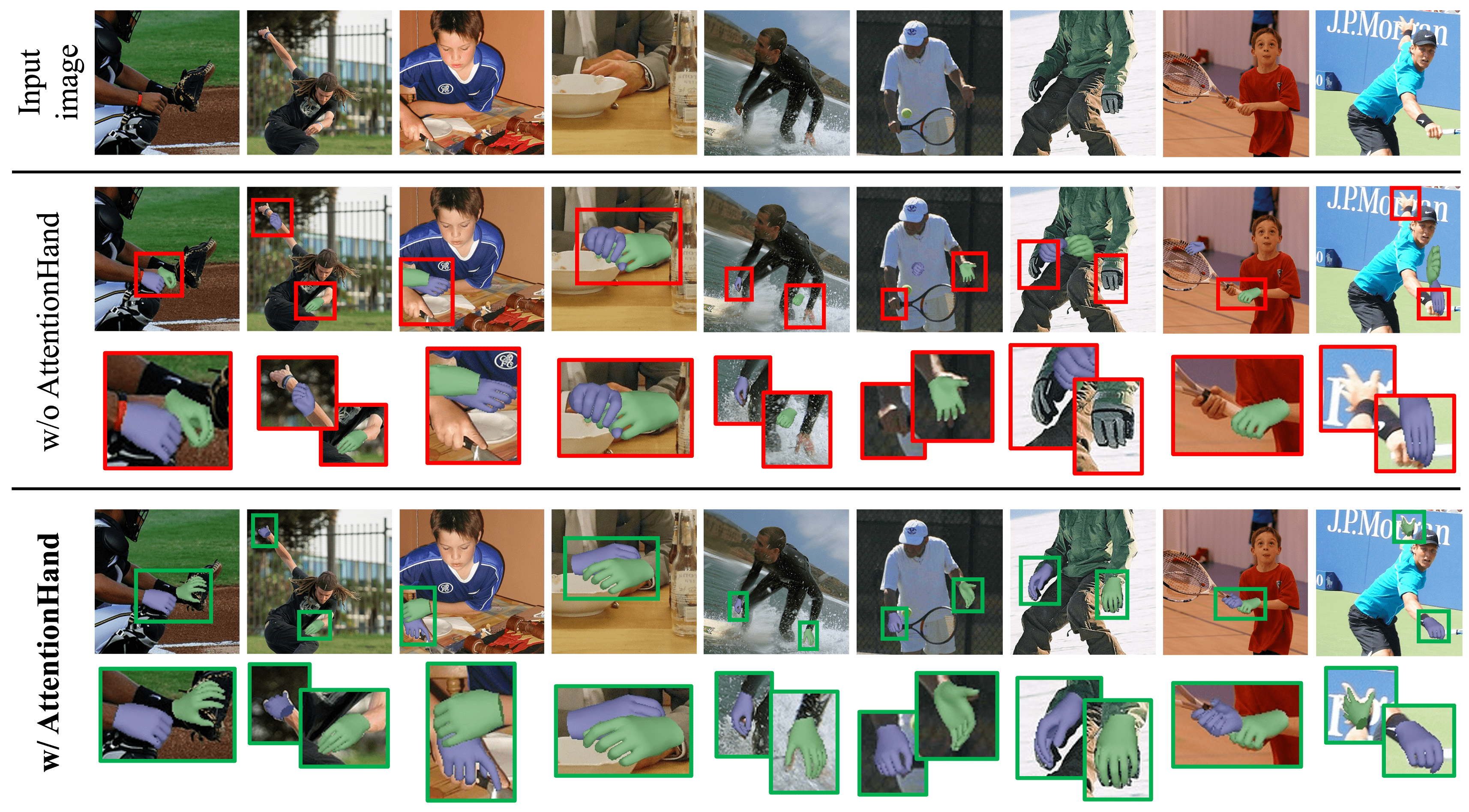}
\end{center}
\caption{Qualitative comparisons on MSCOCO \cite{lin2014microsoft}. 
Red and green boxes indicate wrong and correct region of the reconstructed hand, respectively.}
\label{exp_mesh}
\end{figure*}

\subsubsection{3D Hand Mesh Reconstruction.}
To verify our AttentionHand extensively, we trained state-of-the-art hand pose networks \cite{rong2021monocular, zhang2021interacting, li2022interacting, ren2023decoupled, moon2023bringing} by additionally adding new data generated by AttentionHand.
As shown in Table \ref{tab:mesh}, the performance of all methods increased for all metrics. 
Specifically, with respect to the MPVPE, AttentionHand showed the dramatic performance improvement with InterWild \cite{moon2023bringing} about 3.66\% and 7.81\% on HIC and ReIH, respectively. 
With respect to the RRVE, it increased by about 1.17\% and 0.65\% on HIC and ReIH, respectively. 
With respect to the MRRPE, it increased by about 6.40\% and 1.47\% on HIC and ReIH, respectively. 
These imply generated hand images help increasing the accuracy of the 3D hand mesh reconstruction.
In addition, the qualitative performance for in-the-wild scenes is also verified as shown in Fig. \ref{exp_mesh}. Although MSCOCO mainly contains in-the-wild situations, 3D hand mesh is reconstructed robustly. It implies that even for difficult situations, the performance of reconstruction can be improved by utilizing AttentionHand. Additional qualitative results are in the supplementary materials.

\begin{figure*}[t!]
\centering
\begin{center}
\includegraphics[width=\linewidth]{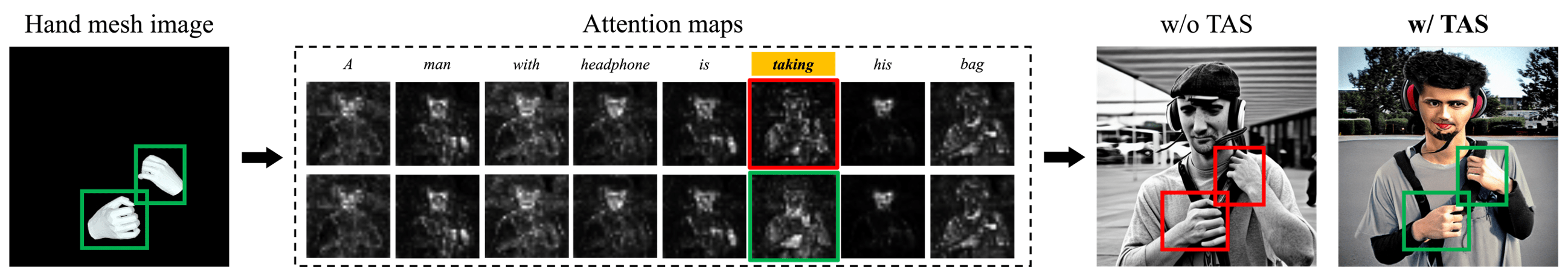}
\end{center}
\caption{Ablation studies on the text attention stage (TAS). 
Attention maps with red and green box are results without and with TAS, respectively.
Red and green bounding boxes indicate wrong and correct hand poses, respectively.
}
\label{exp_TAS}
\end{figure*}

\begin{figure}[t]
\centering
\begin{center}
\includegraphics[width=\linewidth]{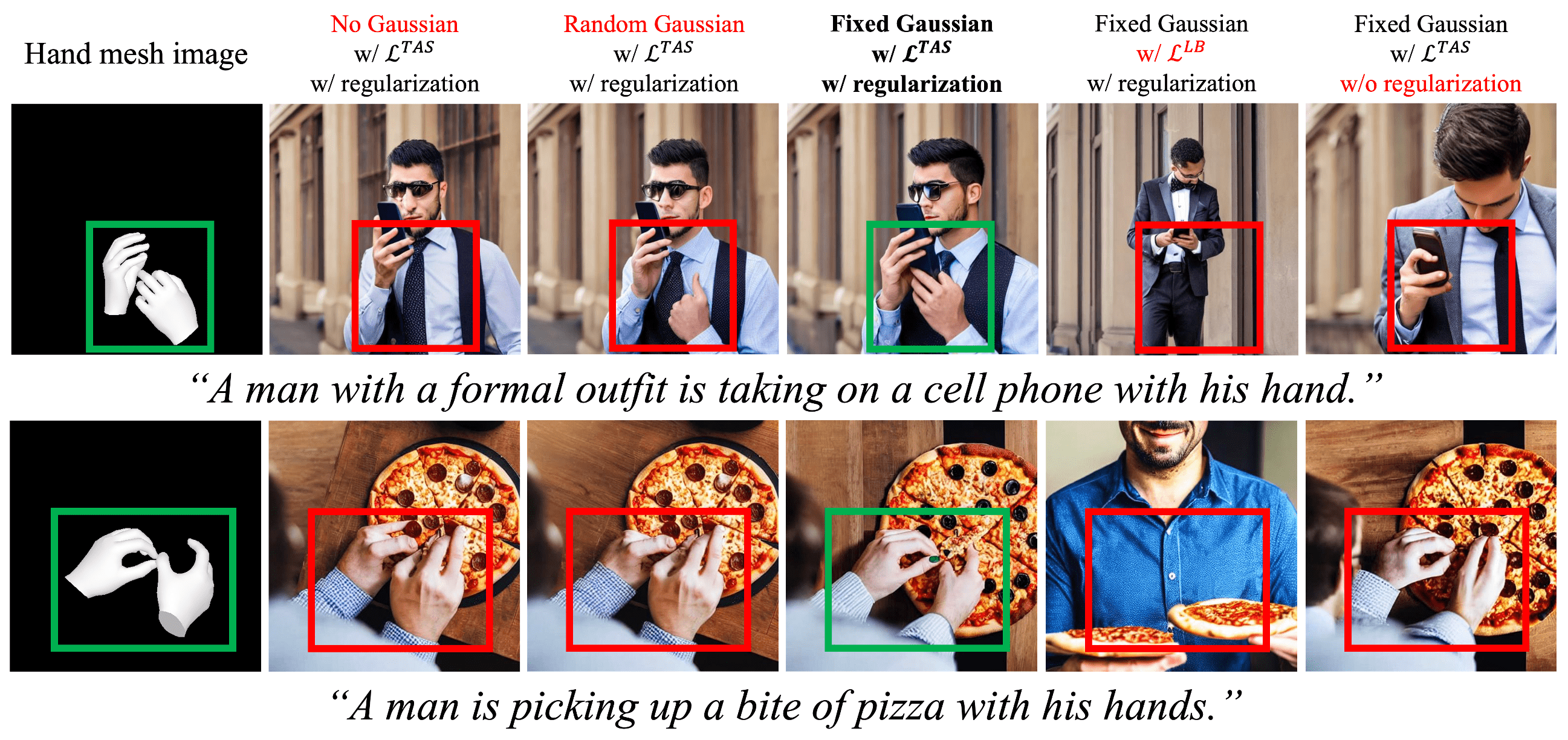}
\end{center}
\caption{Ablation studies on Gaussian filter, losses (i.e. $\mathcal{L}^{TAS}$ and $\mathcal{L}^{LB}$), and the regularization of $\hat{\epsilon}$ for the text attention stage (TAS). Red and green bounding boxes indicate wrong and correct hand poses, respectively.}
\label{exp_Gaussian}
\end{figure}

\subsection{Ablation Studies}

\subsubsection{Text Attention Stage (TAS).}
We deeply dived into TAS to verify its superiority.
Firstly, as in the last two rows in Table \ref{tab:gen}, TAS showed its effectiveness in all metrics. 
In addition, as shown in Fig. \ref{exp_TAS}, attention maps are well described their corresponding tokens in the case of with TAS. 
It implies that with TAS, AttentionHand can reflect hand-related tokens enough.
Additional qualitative results are in the supplementary materials.

Secondly, we conducted more experiments about Gaussian filters as follows: (1) no Gaussian filter, (2) random Gaussian filter, and (3) fixed Gaussian filter. As shown in the second, third, and fourth columns of Fig. \ref{exp_Gaussian}, we found interesting results: in the case of (1), the hand was disappeared or its shape became strange. In the case of (2), generated images are not well-aligned with given hand mesh images. However, in the case of (3), generated images are well-aligned with given hand mesh images and look natural. Hence, we determined fixed Gaussian filter makes the generated image plausibly regardless of diffusion timestep $t$.

Thirdly, we compared our loss, $\mathcal{L}^{TAS}$, with the load balancing loss ($\mathcal{L}^{LB}$) \cite{zhou2022mixture, fedus2022switch}. Since $\mathcal{L}^{LB}$ is an auxiliary loss for balancing loads among experts, it plays a similar role with $\mathcal{L}^{TAS}$, which evenly reflects the image features of all the attention maps. Therefore, we replaced $\mathcal{L}^{TAS}$ to $\mathcal{L}^{LB}$ and considered its feasibility as shown in the fifth column of Fig. \ref{exp_Gaussian}. Unfortunately, in the case of $\mathcal{L}^{LB}$, generated images are not fit at all with given hand mesh images. We guess while $\mathcal{L}^{TAS}$ updates the image embedding based on the spatial information of the attention map, $\mathcal{L}^{LB}$ flattens the 2D attention map as 1D representation, leading to distort spatial knowledge.

Last but not least, we explored the range of updated noise ($\hat{\epsilon}$). According to \cite{chefer2023attend}, we set $\alpha_t$ of Eq. \ref{eqn:TAS} as gradually decreasing according to timestep $t$ (i.e., from $20$ to $10$) for regularization of $\hat{\epsilon}$. However, if $\alpha_t$ is randomly set, $\hat{\epsilon}$ tends to be out of distribution (i.e., Gaussian distribution) as shown in the sixth column of Fig. \ref{exp_Gaussian}: in the case of w/o regularization, generated images are not aligned with given mesh images, or missed some hands. Therefore, it is necessary to regularize $\hat{\epsilon}$ for faithful hand image generation.

\subsubsection{Model Design Justification.}
To justify our model's superiority, we compared the characteristics of prior works including our model.
As shown in Table \ref{tab:network}, our model's distinctive and potential features compared to prior works are (1) harmonious preservation of locality (i.e., hand) with globality (i.e., in-the-wild scene), and (2) selective attention on hand-related tokens by cross attention.
Specifically, to harmonize globality and locality, we developed global and local designs for the visual attention stage (VAS).
Moreover, since the global and local branches are designed structurally same, we set them to share their weights for reducing the number of training parameters (about 20.2\% $\downarrow$) and improving the generalizability (see two shaded rows in Table \ref{tab:VAS}).
We experimentally verified the effectiveness of our design as shown in Table \ref{tab:VAS}.

\subsubsection{Robustness of Generated Dataset.}
\begin{figure*}[t!]
\centering
\begin{center}
\includegraphics[width=\linewidth]{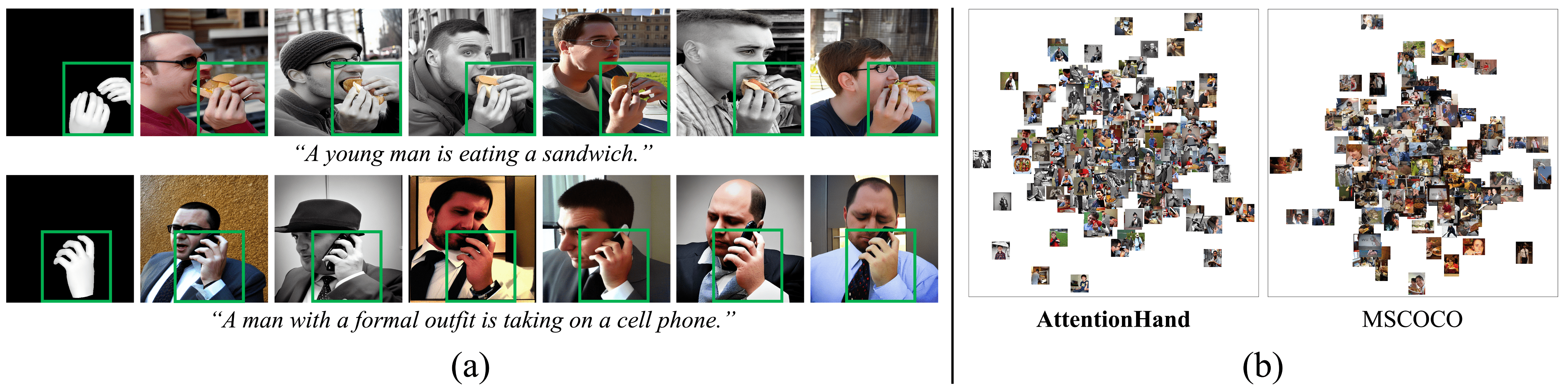}
\end{center}
\caption{(a) Multiple generated hand images from same modalities. Green boxes indicate correct hand poses. (b) t-SNE distribution of AttentionHand and MSCOCO \cite{lin2014microsoft}. }
\label{exp_multi}
\end{figure*}

To verify robustness of our generated dataset, we generated multiple hand images from same modalities as shown in Fig. \ref{exp_multi}(a). As a result, all generated images are perfectly well-aligned with given hand mesh images. Moreover, we found the t-SNE distribution \cite{van2008visualizing} of AttentionHand is broader than MSCOCO as shown in Fig. \ref{exp_multi}(b).
As a result, we believe that AttentionHand can contribute to the downstream task with our extensive in-the-wild hand images, leading to alleviate the domain gap between indoor and outdoor scenes.

\begin{table}[t!]
\caption{Network comparisons with prior works.}
\label{tab:network}
\centering
\resizebox{0.7\linewidth}{!}{
\begin{tabular}{l|c|c|c|c}
\toprule
{Methods} & {Text Prompt} & {Visual Prompt} & {Locality} & {Hand-related Token Attention} \\
\hline
\hline
{Stable Diffusion} & {\cmark} & {} & {} & {} \\
{Uni-ControlNet} & {\cmark} & {\cmark} & {} & {} \\
{T2I-Adapter} & {\cmark} & {\cmark} & {} & {} \\
{ControlNet} & {\cmark} & {\cmark} & {} & {} \\
\rowcolor{Gray}
\textbf{AttentionHand (Ours)} & {\cmark} & {\cmark} & {\cmark} & {\cmark} \\
\bottomrule
\end{tabular}
}
\end{table}

\begin{table}[t!]
\caption{Ablation studies on the visual attention stage (VAS).}
\label{tab:VAS}
\centering
\resizebox{0.7\linewidth}{!}{
\begin{tabular}{ccc|ccccccc}
\toprule
{Globality} & {Locality} & {Weights} & {FID$\downarrow$} & {KID$\downarrow$} & \textcolor{black}{FID-H$\downarrow$} & \textcolor{black}{KID-H$\downarrow$} & \textcolor{black}{Hand Conf.$\uparrow$} & \textcolor{black}{MSE-2D$\downarrow$} & {MSE-3D$\downarrow$} \\
\hline
\hline
{} & {} & {Shared} & {40.52} & {0.00684} & \textcolor{black}{50.78} & \textcolor{black}{0.02554} & \textcolor{black}{0.651} & \textcolor{black}{2.932} & {4.591} \\
{\cmark} & {} & {Shared} & {21.67} & {0.00658} & \textcolor{black}{40.32} & \textcolor{black}{0.02098} & \textcolor{black}{0.810} & \textcolor{black}{1.252} & {2.182} \\
{} & {\cmark} & {Shared} & {52.98} & {0.00713} & \textcolor{black}{32.11} & \textcolor{black}{0.01604} & \textcolor{black}{0.911} & \textcolor{black}{1.539} & {2.397} \\
\rowcolor{Gray}
{\cmark} & {\cmark} & \textbf{Shared} & \textbf{20.71} & {0.00301} & {\textcolor{black}{27.09}} & \textbf{\textcolor{black}{0.01287}} & \textbf{\textcolor{black}{0.965}} & \textbf{\textcolor{black}{1.026}} & \textbf{1.986} \\
\rowcolor{Gray2}
{\cmark} & {\cmark} & {Separated} & {21.90} & \textbf{0.00293} & \textbf{\textcolor{black}{26.89}} & \textcolor{black}{0.01340} & \textcolor{black}{0.960} & \textcolor{black}{1.108} & {2.017} \\
\bottomrule
\end{tabular}
}
\end{table}

\section{Conclusion}

In this paper, we introduced a novel text-to-hand image generation model, AttentionHand, which pays attention to the hand-related tokens from the text prompt and global and local mesh images.
AttentionHand achieved state-of-the-art performance in text-to-hand image generation, and we demonstrated that training with the dataset generated by our AttentionHand improved the performance of 3D hand mesh reconstruction. 
However, the diversity may decrease as the generative model is trained to optimize hand mesh images. We expect for the emergence of outstanding diffusion model to improve the diversity and quality of the hand image.

\noindent \textbf{Acknowledgements.} This research was supported by the MSIT (Ministry of Science and ICT), Korea, under the ITRC (Information Technology Research Center) support program (IITP-2024-RS-2023-00260091) supervised by the IITP (Institute for Information \& Communications Technology Planning \& Evaluation) and Korea Institute for Advancement of Technology (KIAT) grant funded by the Korea Government (MOTIE) (P0020535, The Competency Development Program for Industry Specialist) and National Supercomputing Center with supercomputing resources including technical support (KSC-2023-CRE-0444).

\title{Supplementary Materials for\\``AttentionHand:\\Text-driven Controllable Hand Image Generation for 3D Hand Reconstruction in the Wild''} 
\titlerunning{AttentionHand}
\author{}
\authorrunning{J. Park et al.}
\institute{}
\maketitle

\renewcommand*{\thesection}{\Alph{section}}
\renewcommand{\thefigure}{\Alph{figure}}
\renewcommand{\thetable}{\Alph{table}}

\section{Preliminary: Latent Diffusion Model}

Latent Diffusion Model (LDM) or Stable Diffusion (SD) \cite{rombach2022high} is a type of diffusion model designed for training in a latent space, which is particularly well-suited for likelihood-based generative models. Unlike traditional approaches that utilize the full high-dimensional pixel space, LDM leverages the latent space to concentrate on the essential and meaningful aspects of the data. This enables training in a lower-dimensional space, resulting in significantly improved computational efficiency. The objective of LDM is as follows:
\begin{equation}
L=\mathbb{E}_{z_0,t,c,\epsilon \sim \mathcal{N}(0,1)}[\|\epsilon-\epsilon_{\theta}(z_{t}, t, c)\|^2_2],
\end{equation}
where $z_0$ is the initial latent image, $z_t$ is the noisy latent image after $t$ diffusion steps of $z_0$, $c$ is the text embedding, and $\epsilon_{\theta}$ is the diffusion training network.

\section{Details of Data Preparation Phase}

To train AttentionHand, it just requires easy-to-use four modalities: (1) a global RGB hand image $I_{RGB}^G\in \mathbb{R}^{3\times512\times512}$, which represents in-the-wild scene with hand, (2) the corresponding global hand mesh image $I^{G}_{mesh} \in \mathbb{R}^{3 \times 512 \times 512}$, (3) the corresponding bounding box of hand region $B \in \mathbb{R}^{1 \times 4}$, and (4) the corresponding hand-related text prompt $U$.

\subsubsection{Rendering Hand Mesh Images.}
$I^{G}_{mesh}$ is obtained by utilizing MANO \cite{romero2017embodied}, which is generally adopted as ground-truth for 3D hand mesh reconstruction. Specifically, the ground-truth root pose $M_{root}\in \mathbb{R}^{1\times3}$, hand pose $M_{hand}\in \mathbb{R}^{15\times3}$, shape $M_{shape}\in \mathbb{R}^{1\times10}$, translation $M_{trans}\in \mathbb{R}^{1\times3}$, and hand type $h$ (i.e., left or right hand) are passed to MANO layer to get the mesh $M_{mesh}\in \mathbb{R}^{778\times3}$ and face $M_{face}\in \mathbb{R}^{1538\times3}$ of 3D hand as following:
\begin{equation}
M_{mesh}, M_{face}=ManoLayer(M_{root},M_{hand},M_{shape},M_{trans},h).
\end{equation}
Then, by using $M_{mesh}$ and $M_{face}$ with ground-truth camera rotation matrix $R\in \mathbb{R}^{3\times3}$, camera translation matrix $t\in \mathbb{R}^{1\times3}$, focal length $f\in \mathbb{R}^{1\times2}$, and principal point $p\in \mathbb{R}^{1\times2}$, we can render the 2D mesh image $I^{G}_{mesh}$ as following:
\begin{equation}
I^{G}_{mesh}=Render(M_{mesh}, M_{face},R,t,f,p),
\end{equation}
where $Render$ is operated based on PyTorch3D \cite{ravi2020accelerating}.

\subsubsection{Captioning with Hand-related Text Prompt.}
\begin{figure*}[t!]
\centering
\begin{center}
\includegraphics[width=\linewidth]{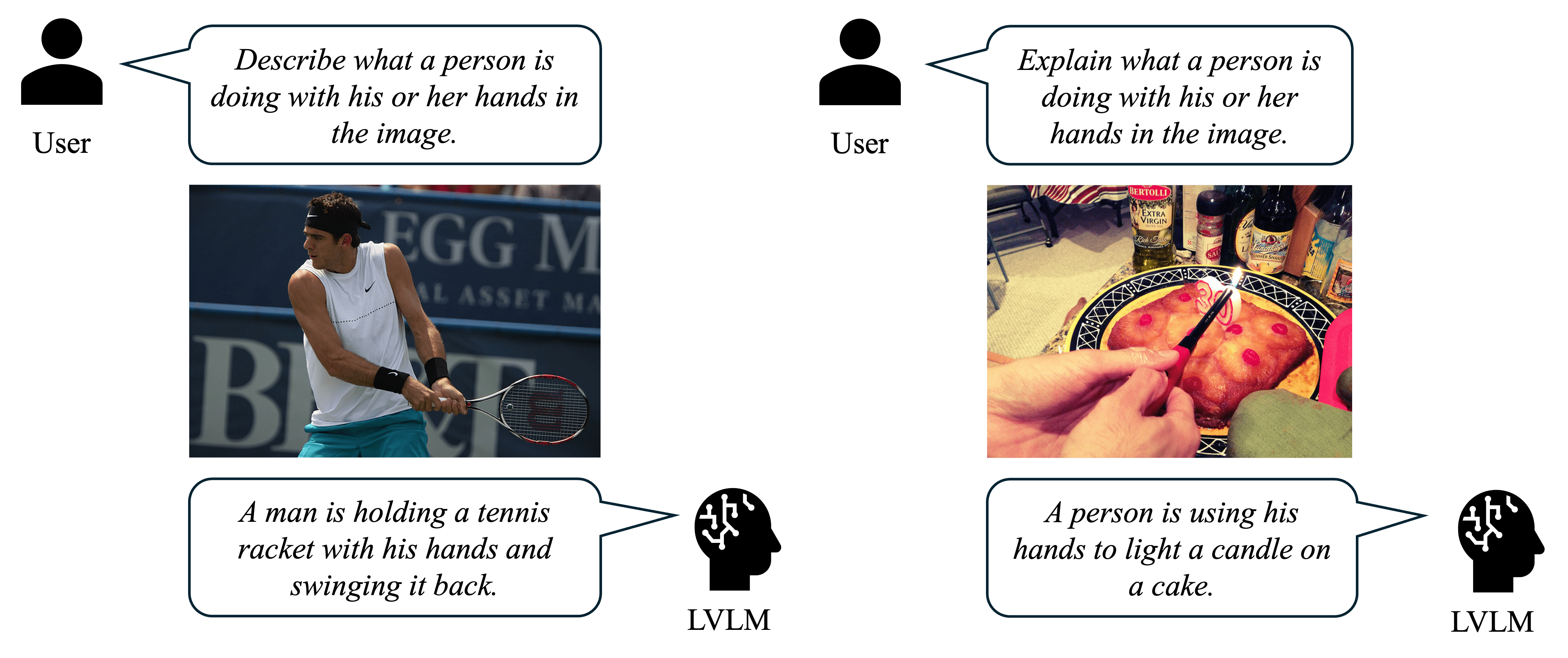}
\end{center}
\caption{Examples of captioning process with the off-the-shelf large vision language model (LVLM).}
\label{supple_captioning}
\end{figure*}

To caption an image with hand-related text prompt $U$, we employ Qwen-VL \cite{bai2023qwen}, the off-the-shelf large vision language model. Specifically, by entering the image and the question such as \textit{``Describe what a person is doing with his or her hands in the image.''} into the Qwen-VL, the answer about what a person does with his or her hand can be obtained. Examples of this process can be found in Fig. \ref{supple_captioning}.

\begin{figure*}[t!]
\centering
\begin{center}
\includegraphics[width=\linewidth]{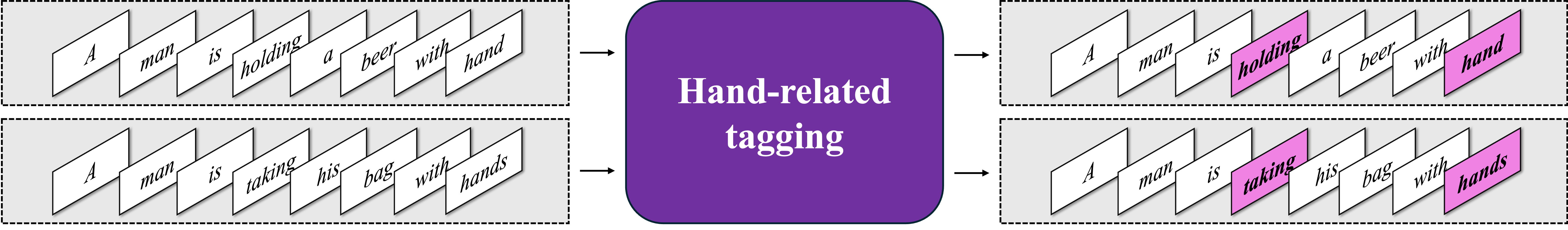}
\end{center}
\caption{Examples of the hand-related tagging process.}
\label{supple_tagging}
\end{figure*}

\section{Hand-related Tagging}
As we mentioned in the section 3.3 of the main body, we design the hand-related tagging $\mathcal{H}_{tag}$, which is based on part-of-speech tagging \cite{chiche2022part} from NLTK library \cite{loper2002nltk}. Specifically, $\mathcal{H}_{tag}$ determines if the input token`s part of speech indicates ``VBG'' (i.e., \textit{holding}, \textit{taking}, \textit{using}, etc), or if the input token contains the word \textit{hand(s)}. As a result, we can extract hand-related tokens with $\mathcal{H}_{tag}$. Examples of this process can be found in Fig. \ref{supple_tagging}.

\section{Details of Experiments}
\subsection{Dataset}
\subsubsection{Text-to-Image Generation.}
For the train, RGB hand images, mesh images, and bounding boxes were utilized from the train set of MSCOCO \cite{lin2014microsoft}. Text prompts, which represent hand-related descriptions, are obtained by off-the-shelf captioning model \cite{bai2023qwen}.
Note that since two or more people's hands can be seen as one person's hands on the hand-focused image, we filtered out the case of more than one person for the data preparation. Hence, AttentionHand is induced to train about single or both hands of only one person. 
For the test, RGB hand images and mesh images were also utilized from the train set of MSCOCO to evaluate the image quality and pose alignment of generated hand images. On the other hand, text prompts were utilized from the validation set of MSCOCO. 
Moreover, we adopted Hands-In-Action (HIC) \cite{tzionas2016capturing}, Re:InterHand (ReIH) \cite{moon2023dataset}, and InterHand2.6M (IH2.6M) \cite{moon2020interhand2} to evaluate the effectiveness of generated hand images for the 3D hand mesh reconstruction.

\subsubsection{3D Hand Mesh Reconstruction.}

For the train, we utilized hand mesh images from ReIH and IH2.6M, which provide accurate 3D hand labels, to generate new training samples.
For the test, we adopted HIC, ReIH, IH2.6M, and MSCOCO to evaluate the accuracy of reconstructed 3D hand mesh.

\subsection{Evaluation Protocol}
\subsubsection{Text-to-Image Generation.}
\begin{figure*}[t!]
\centering
\begin{center}
\includegraphics[width=0.7\linewidth]{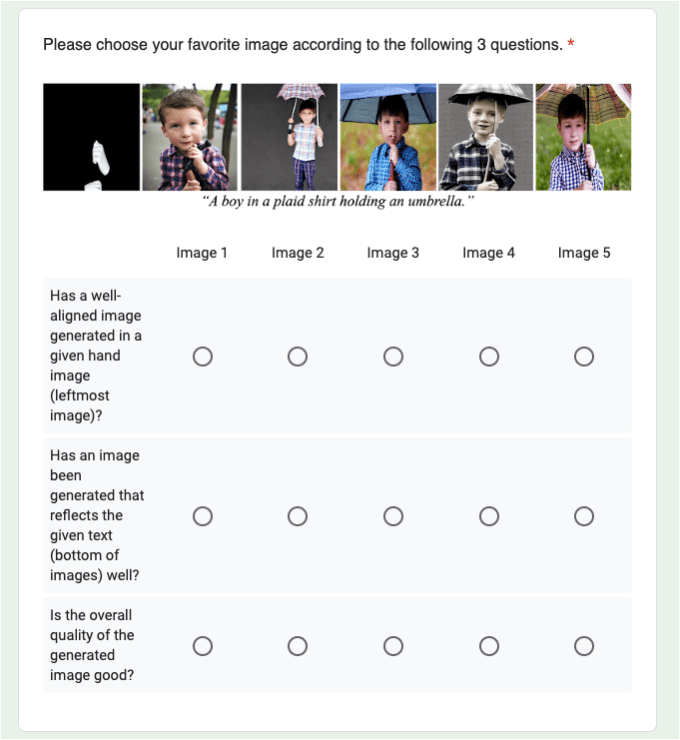}
\end{center}
\caption{Screenshot of our user preference. For each user study, only one sample among 24 samples is shown.}
\label{supple_user}
\end{figure*}

To evaluate the image quality, we adopted frechet inception distance (FID) \cite{heusel2017gans} and kernel inception distance (KID) \cite{binkowski2018demystifying}.
In addition, according to \cite{narasimhaswamy2024handiffuser}, we computed FID-Hand (FID-H), KID-Hand (KID-Hand), and the hand confidence score (Hand Conf.), to measure the quality of images only in the hand regions. 
To evaluate the pose alignment, we adopted the mean square error of 3D keypoints (MSE-3D) for analysis the error between the ground-truth and predicted keypoints estimated by the off-the-shelf model \cite{moon2023bringing}.
Additionally, to validate reliability, we evaluated the mean square error of 2D keypoints (MSE-2D) using Mediapipe \cite{zhang2020mediapipe}.
Moreover, similar to \cite{rombach2022high, ye2023affordance}, we carried out user preference to evaluate the perceptual plausibility of generated images. Specifically, we attached 24 samples of results in the Google Forms, and released it to 30 people. We asked for three questions as shown in Fig. \ref{supple_user}: (1) alignment with the given mesh image, (2) reflection of the given text prompt, and (3) overall quality of the generated image. The results of these questions were averaged and quantified in percentage.

\subsubsection{3D Hand Mesh Reconstruction.}
We adopted the mean per-vertex position error (MPVPE), the right hand-relative vertex error (RRVE), and the mean relative-root position error (MRRPE), which are representative metrics for the 3D hand mesh reconstruction.

\subsection{Implementation Details}
\subsubsection{Text-to-Image Generation.}
For the text-to-image generation, we adopted PyTorch Lightning \cite{Lightning-AI} framework. We set the batch size as 1, and learning rate as $10^{-5}$. We used one RTX 3090.

\subsubsection{3D Hand Mesh Reconstruction.}
For the 3D hand mesh reconstruction, we mainly referred to InterWild \cite{moon2023bringing}. Specifically, we adopted PyTorch \cite{paszke2017automatic} framework. We set the batch size as 32, and learning rate as $10^{-4}$ for the first 4 epochs, as $10^{-5}$ for the rest epochs. We used one RTX 3090.

\subsection{Generalizability of AttentionHand}

To verify the generalizability of AttentionHand for 3D hand mesh reconstruction, we additionally generated hand images with state-of-the-arts of text-to-image generation, utilized them as training sets of the off-the-shelf model \cite{moon2023bringing}, which is suitable for in-the-wild generalization, and tested on in-the-wild datasets (i.e., HIC and ReIH) and in-the-lab dataset (i.e., IH2.6M.) As a result, AttentionHand achieved the highest performance in all test sets as shown in Table \ref{tab:supple_gen}. 
It implies that since AttentionHand makes various and well-aligned in-the-wild images, it enhances the robustness of the 3D hand mesh reconstruction model.

\begin{table}[t!]
\caption{Quantitative comparisons on the 3D hand mesh reconstruction with state-of-the-art text-to-image generation models.}
\label{tab:supple_gen}
\centering
\resizebox{1.0\textwidth}{!}{
\begin{tabular}{l|ccc|ccc|ccc}
\toprule
\multicolumn{1}{r|}{Datasets} & \multicolumn{6}{c|}{In-the-wild} & \multicolumn{3}{c}{In-the-lab} \\
\cline{2-10}
 & \multicolumn{3}{c|}{HIC \cite{tzionas2016capturing}} & \multicolumn{3}{c|}{ReIH \cite{moon2023dataset}} & \multicolumn{3}{c}{IH2.6M \cite{moon2020interhand2}} \\
\cline{2-10}
\multicolumn{1}{l|}{Methods} & {MPVPE$\downarrow$} & {RRVE$\downarrow$} & {MRRPE$\downarrow$} &  {MPVPE$\downarrow$} & {RRVE$\downarrow$} & {MRRPE$\downarrow$} & {MPVPE$\downarrow$} & {RRVE$\downarrow$} & {MRRPE$\downarrow$} \\
\hline
\hline
Stable Diffusion \cite{rombach2022high} & 16.97 & 32.33 & 41.11 & 18.78 & 29.00 & 32.09 & 14.03 & 25.70 & 33.95  \\
Uni-ControlNet \cite{zhao2023uni} & 16.19 & 27.03 & 32.10 & 17.08 & 24.91 & 27.23 & 12.01 & 22.12 & 31.08  \\
T2I-Adapter \cite{mou2023t2i} & 16.06 & 25.67 & 36.53 & 16.99 & 25.87 & 29.88 & 12.20 & 21.48 & 30.93  \\
ControlNet \cite{zhang2023adding} & 15.43 & 24.11 & 30.75 & 15.12 & 23.60 & 26.17 & 11.53 & 20.99 & 26.24  \\
{AttentionHand (w/o TAS)} & 14.85 & 22.47 & 29.99 & 14.65 & 21.57 & 25.89 & 10.75 & 20.89 & 26.02  \\
\rowcolor{Gray}
\textbf{AttentionHand (w/ TAS)} & \textbf{14.74} & \textbf{21.10} & \textbf{29.26} & \textbf{13.95} & \textbf{19.94} & \textbf{22.05} & \textbf{10.62} & \textbf{19.09} & \textbf{25.74}  \\
\bottomrule
\end{tabular}
}
\end{table}

\subsection{More Ablation Study of Text Attention Stage}

\begin{figure*}[t!]
\centering
\begin{center}
\includegraphics[width=\linewidth]{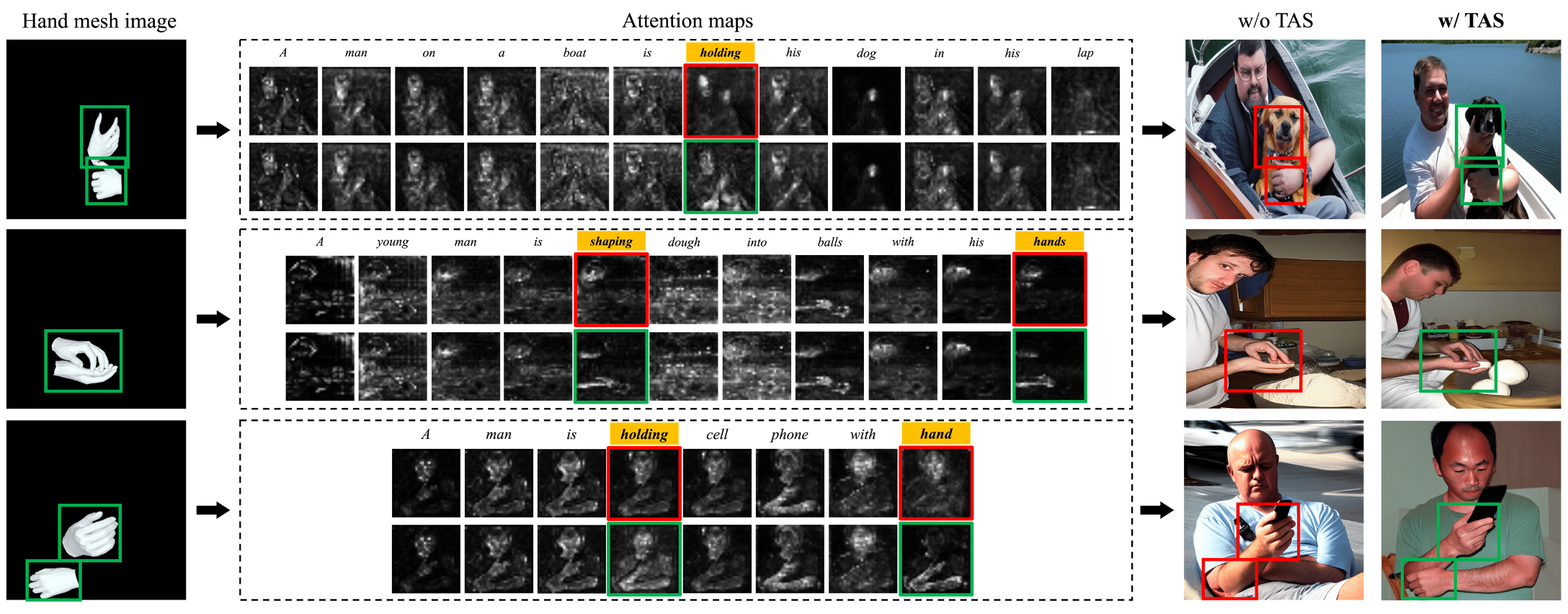}
\end{center}
\caption{
Ablation studies on the text attention stage (TAS). 
Attention maps with red and green box are results without and with TAS, respectively.
Red and green bounding boxes indicate wrong and correct hand poses, respectively.
}
\label{supple_exp_TAS}
\end{figure*}

We additionally verified the effectiveness of the text attention stage (TAS) as shown in Fig. \ref{supple_exp_TAS}. Specifically, based on given hand mesh images and text prompts, we visualized attention maps and generated new hand images with three cases. Without TAS, attention of corresponding tokens was not well represented as shown in attention maps with red boxes. However, with TAS, attention was more highlighted by reflecting corresponding tokens as shown in attention maps with green boxes. It implies that with TAS, AttentionHand can reflect hand-related tokens enough. 

\subsection{More Qualitative Results}
\subsubsection{Text-to-Image Generation.}
\begin{figure*}[t!]
\centering
\begin{center}
\includegraphics[width=\linewidth]{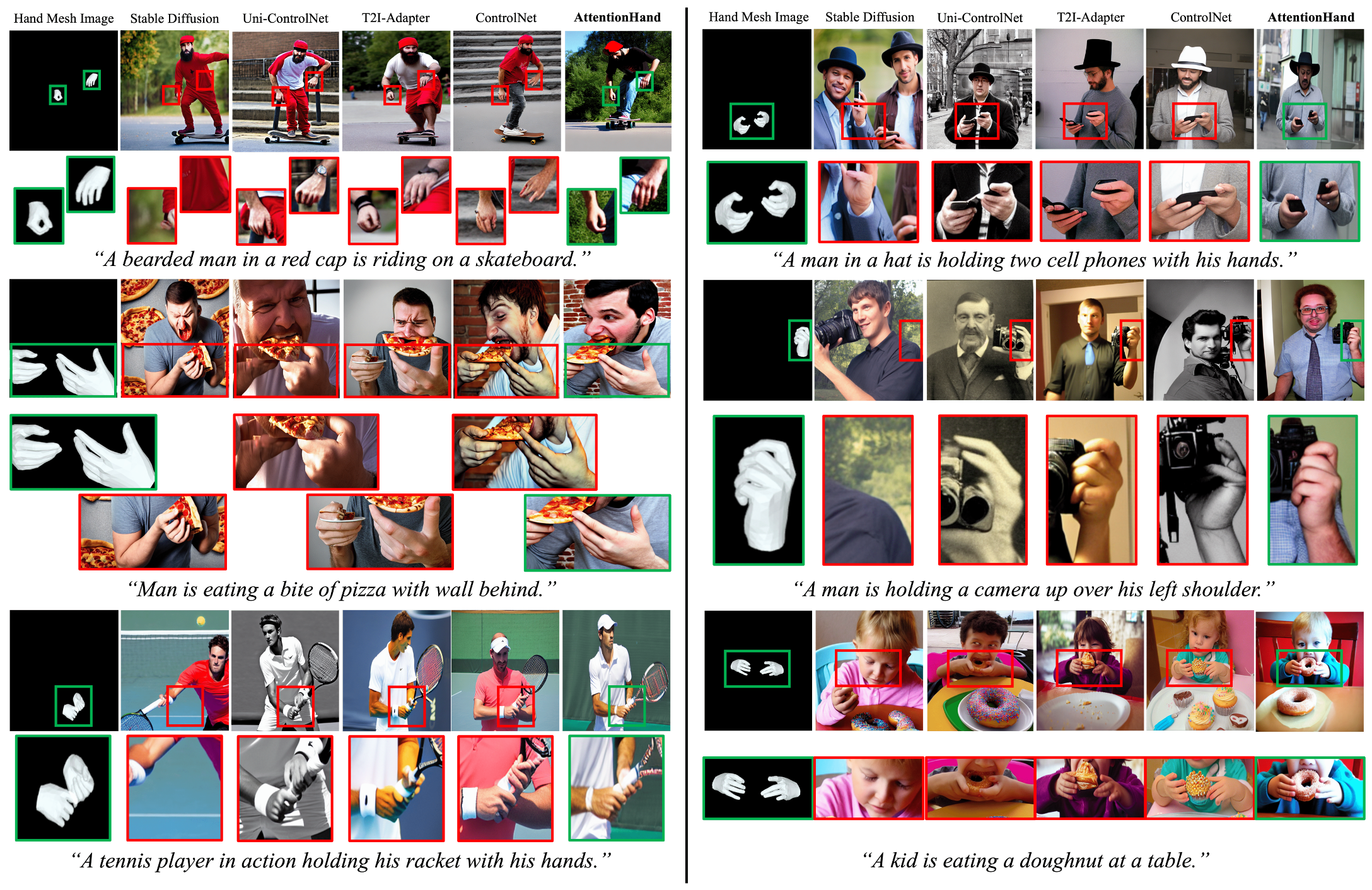}
\end{center}
\vspace{-0.5cm}
\caption{Qualitative comparisons with state-of-the-art text-to-image generation models.
Red and green boxes in each sample indicate the wrong and correct hand bounding box, respectively.}
\label{supple_exp_gen1}
\end{figure*}

\begin{figure*}[t!]
\centering
\begin{center}
\includegraphics[width=\linewidth]{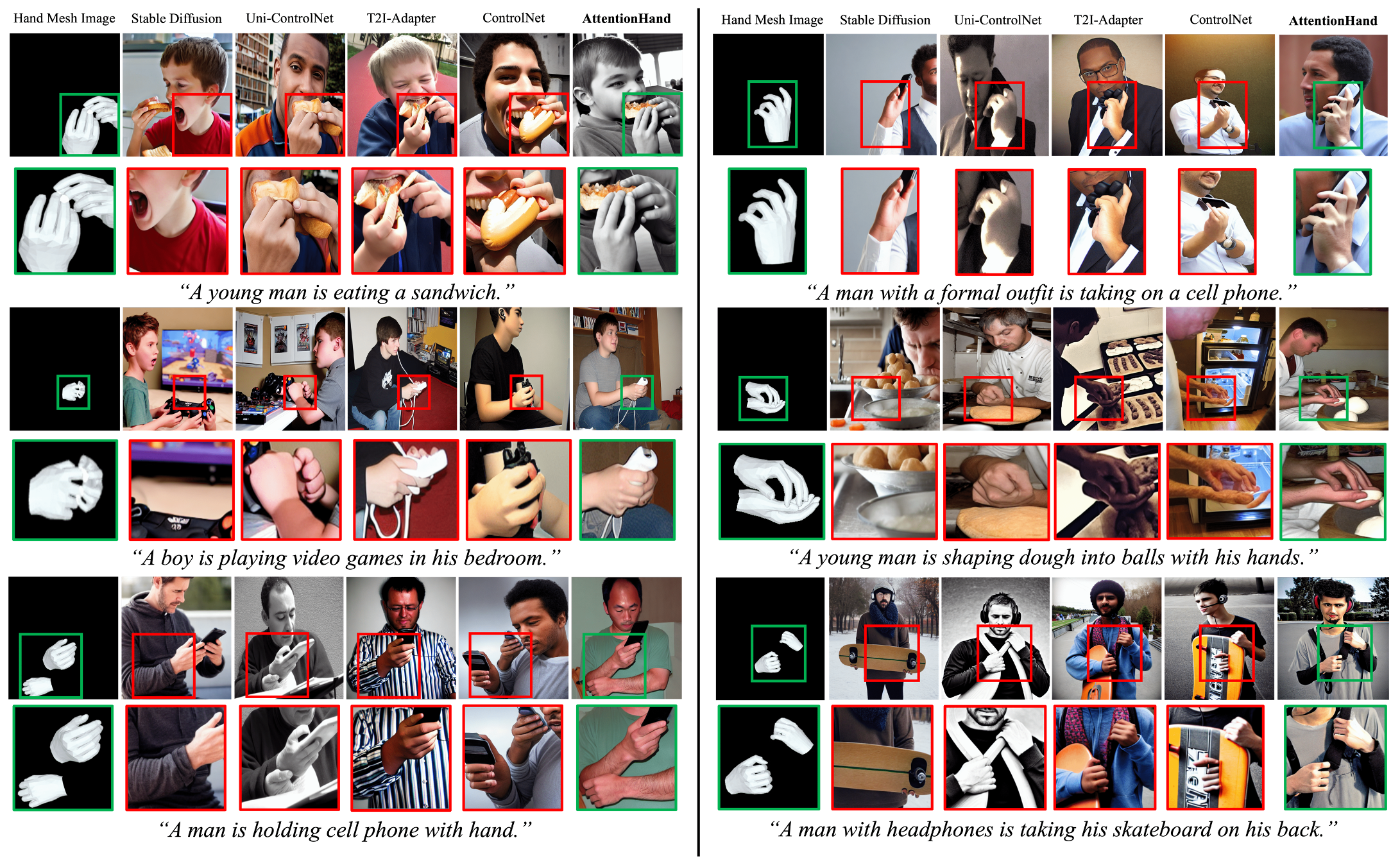}
\end{center}
\vspace{-0.5cm}
\caption{Qualitative comparisons with state-of-the-art text-to-image generation models.
Red and green boxes in each sample indicate the wrong and correct hand bounding box, respectively.}
\label{supple_exp_gen2}
\end{figure*}

Compared to state-of-the-arts \cite{rombach2022high, zhao2023uni, mou2023t2i, zhang2023adding}, our AttentionHand generated the high-quality hand image which is well-aligned with the given mesh image and fully reflected the given text prompt as shown in Figs. \ref{supple_exp_gen1} and \ref{supple_exp_gen2}. Specifically, even when two-hands mesh image is given, which is more challenging than in the case of single-hand mesh image, AttentionHand generated the hand image robustly. It implies our AttentionHand is proper to generate well-aligned hand images with given mesh images and text prompt.

\subsubsection{3D Hand Mesh Reconstruction.}

\begin{figure*}[t!]
\centering
\begin{center}
\includegraphics[width=\linewidth]{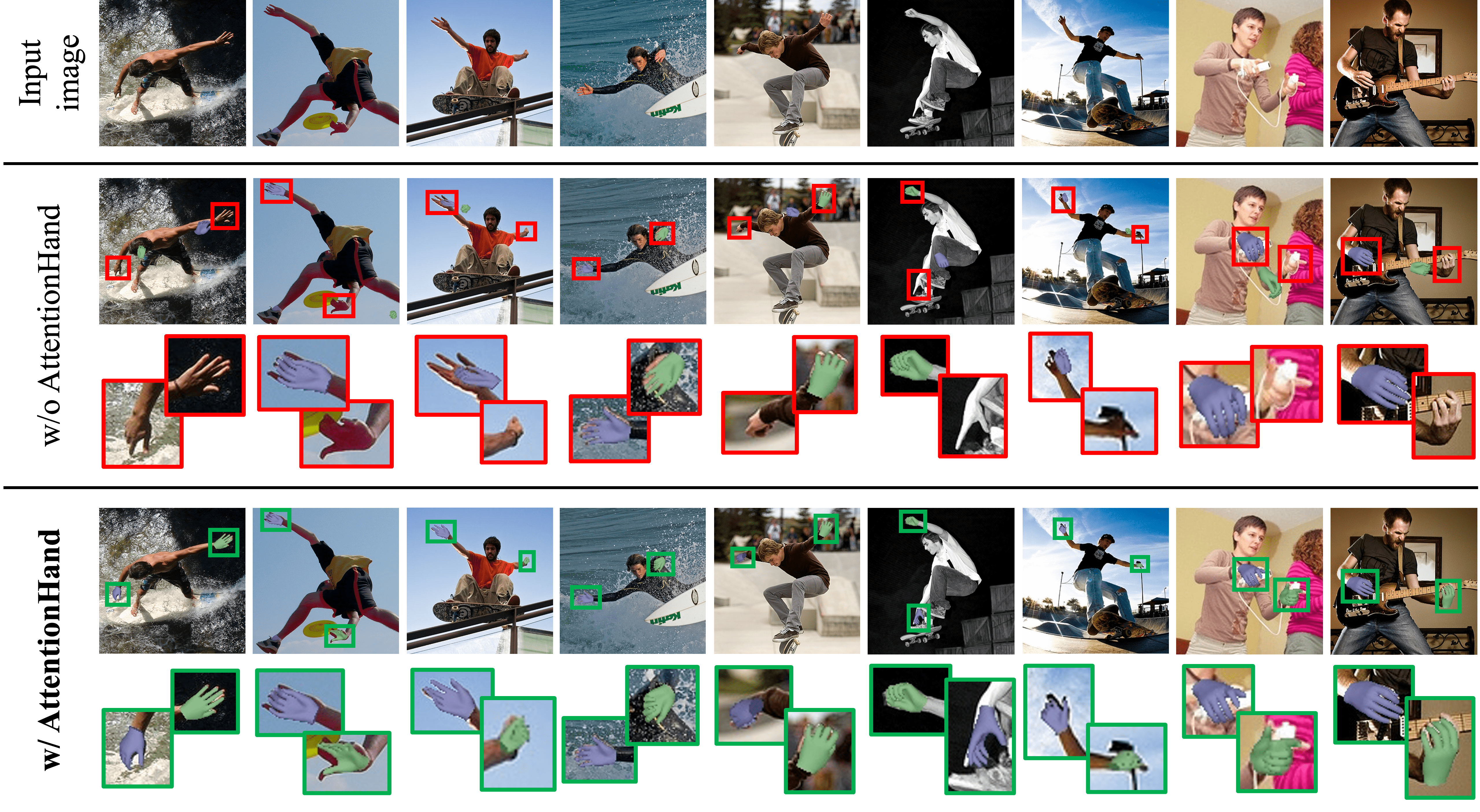}
\end{center}
\caption{Qualitative comparisons on MSCOCO \cite{lin2014microsoft}. 
Red and green boxes indicate wrong and correct region of the reconstructed hand, respectively.}
\label{supple_exp_mesh1}
\vspace{-0.5cm}
\end{figure*}

\begin{figure*}[t!]
\centering
\begin{center}
\includegraphics[width=\linewidth]{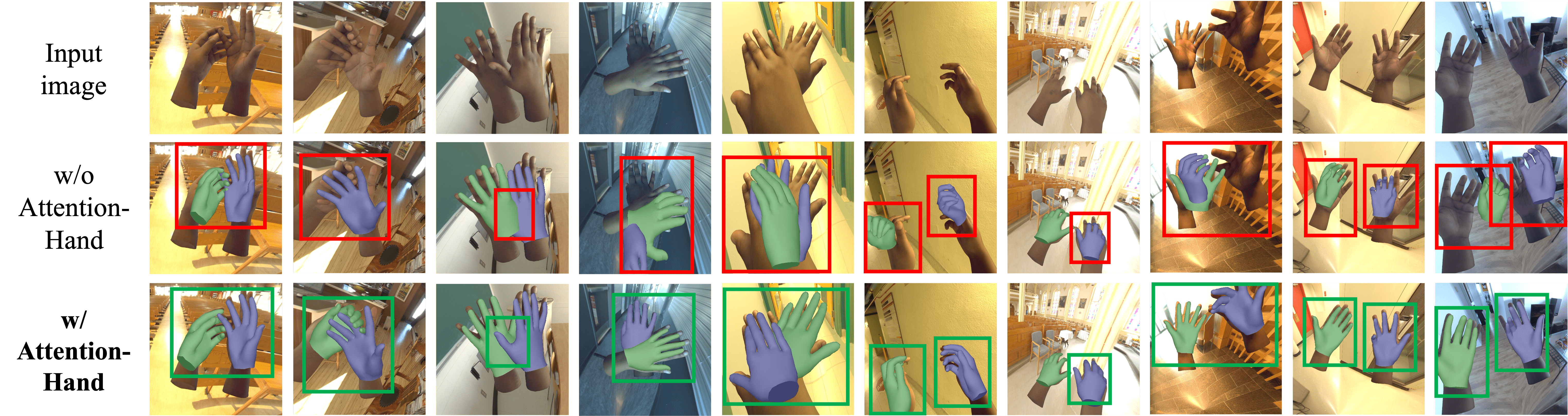}
\end{center}
\caption{Qualitative comparisons on egocentric hands of ReIH \cite{moon2023dataset}. 
Red and green boxes indicate wrong and correct region of the reconstructed hand, respectively.}
\label{supple_exp_mesh2}
\vspace{-0.5cm}
\end{figure*}

\begin{figure*}[t!]
\centering
\begin{center}
\includegraphics[width=\linewidth]{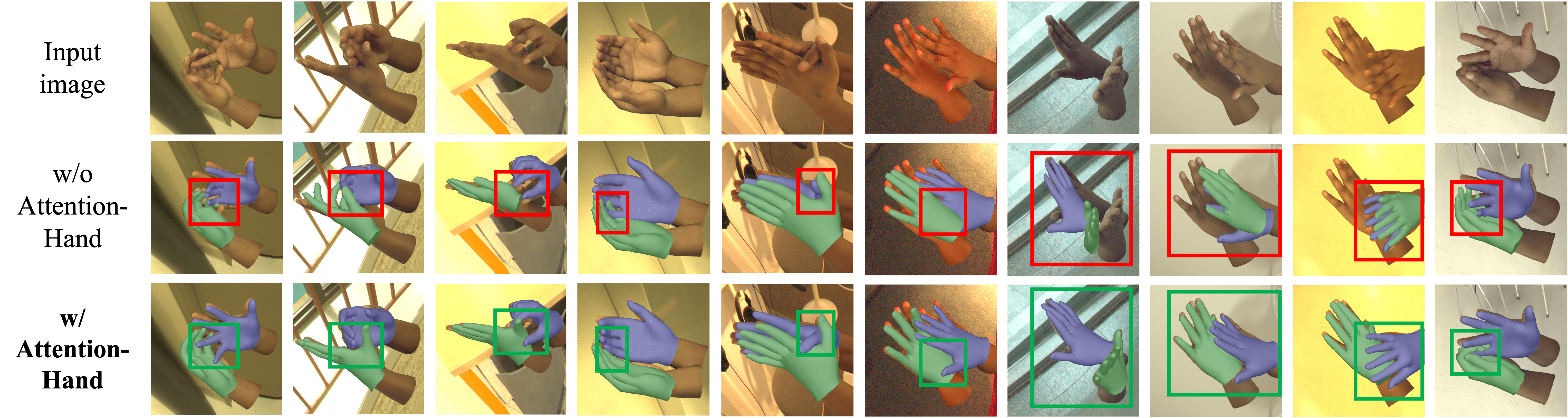}
\end{center}
\caption{Qualitative comparisons on exocentric hands of ReIH \cite{moon2023dataset}. 
Red and green boxes indicate wrong and correct region of the reconstructed hand, respectively.}
\label{supple_exp_mesh3}
\vspace{-0.5cm}
\end{figure*}

We trained off-the-shelf model \cite{moon2023bringing} by additionally adding new data generated by AttentionHand, and tested on MSCOCO and ReIH.
The performance for in-the-wild scenes is verified as shown in Fig. \ref{supple_exp_mesh1}. Although MSCOCO mainly contains in-the-wild situations, 3D hand mesh is reconstructed robustly. It implies that even for difficult situations, the performance of reconstruction can be improved by utilizing AttentionHand.
In addition, the performance improvement is also verified as shown in Figs. \ref{supple_exp_mesh2} and \ref{supple_exp_mesh3}. Note that ReIH is considered more challenging than other datasets because it consists of images with various backgrounds and complex interacting hands. However, by employing AttentionHand, the 3D hand mesh was reconstructed accurately regardless of the viewpoint (i.e., egocentric and exocentric view.) In addition, both interacting hands were elaborately recovered even when hands are in self-handed occlusion and depth ambiguity.

\clearpage

\vspace{-0.5cm}

\clearpage

\vspace{-0.5cm}

\vspace{-0.5cm}

\clearpage


%
%
\bibliographystyle{unsrt}

\end{document}